\documentclass{article}

\usepackage[main,final]{neurips_2026}
\usepackage{amsmath, mathtools, stmaryrd}

\usepackage[utf8]{inputenc} 
\usepackage[T1]{fontenc}    
\usepackage{hyperref}       
\hypersetup{
  pdftitle={Spectral Degeneration of Self-Attention under Isotropic Score Geometry},
  pdfauthor={Mengda Li and Jianfeng Yao}}
\usepackage{url}            
\usepackage{booktabs}       
\usepackage{amsfonts}       
\usepackage{nicefrac}       
\usepackage{microtype}      
\usepackage{xcolor}         
\usepackage[inkscapelatex=false,inkscapearea=page]{svg}
\usepackage{placeins}

\usepackage{amsthm}

\newtheorem{assumption}{Assumption}
\newtheorem{theorem}{Theorem}

\newtheorem{lemma}{Lemma}
\newtheorem{remark}{Remark}
\newtheorem{definition}{Definition}
\DeclareMathOperator{\srank}{srank}
\DeclareMathOperator{\erank}{erank}

\title{Spectral Degeneration of Self-Attention \\ under Isotropic Score Geometry}

\author{%
  Mengda Li and Jianfeng Yao\\
  The Chinese University of Hong Kong, Shenzhen\\
  \texttt{jeffyao@cuhk.edu.cn} \\
}

\begin{document}

\maketitle

\begin{abstract}
We study the spectral structure of the self-attention matrix
in Transformer models
through an idealized spectral model for the pre-softmax score matrix.
Motivated by the rotational isotropy of query and key representations, our model replaces the random token-space orientation of the score matrix by its canonical singular-value representative with a logarithmic spike--bulk separation.
Empirical diagnostics on pretrained vision and language transformers motivate this model: token-space singular matrices have nearly uniform eigen-angle distributions, and score spectra are dominated by a small number of effective directions.
We then prove that the row-wise softmax map transforms this structure into
a degenerate attention matrix: dominant modes are preserved, whereas the bulk spectral distributions concentrate at zero.
Stable rank remains bounded for bidirectional attention but grows logarithmically for causal attention. Within the canonical model, these results explain how attention can preserve a small number of dominant directions while compressing the remaining bulk modes.
\end{abstract}

\section{Introduction}

Transformers have become the dominant architecture in modern machine learning, powering large language models and state-of-the-art systems in vision, speech, and multimodal learning.
Introduced in the seminal work \emph{``Attention Is All You Need''} \citep{vaswani2017attention},
the transformer replaces recurrence and convolution with a simple yet powerful mechanism: \emph{self-attention}.

Given query and key matrices $Q, K$, self-attention computes a score matrix
\(
S = QK^\top / \sqrt{d_{qk}},
\)
followed by a row-wise softmax normalization
\(
A = \mathrm{softmax}(S),
\)
which defines how information is aggregated across tokens.
This mechanism is at the core of language models such as BERT \citep{devlin-etal-2019-bert, liu2019RoBERTa} and GPT-style architectures \citep{GPT1,GPT2,GPT3}
and vision models such as Vision Transformers (ViT) \citep{dosovitskiy_vit_2021}, enabling scalable modeling of long-range dependencies.

Despite its empirical success, the theoretical understanding of attention remains limited,
especially in the \emph{trained regime}.
A central open question is to characterize the structure of the attention matrix $A$,
which governs information flow and expressivity \citep{bordelon2024infinite, sakai2026infinitewidth}.
Recent studies \citep{attention_is_not_depth, anagnostidis2022signal, mindthegap_width, hayase2026gaussian, sanyal2026tmlr} suggest that attention matrices often exhibit
\emph{degeneration phenomena}, where they become close to low-rank or even rank-one structures in deep layers.
Understanding the origin of this degeneration is critical for explaining
both the strengths and limitations of transformer models.

In this paper, we propose a new perspective based on the geometry of the query and key matrices.
Motivated by isotropy diagnostics in trained representations, we study a canonical spectral model that isolates a mechanism of attention degeneration under softmax normalization.
Specifically, we hypothesize that in trained transformers, the query and key matrices exhibit an approximate \emph{token-space orthogonal invariance}, meaning that their token-space singular vectors behave as if they were drawn from a Haar distribution.

This is analogous to the \emph{feature-space rotational invariance} exploited in RaBitQ \citep{RaBitQ24,RaBitQ25} and TurboQuant \citep{zandieh2026turboquant}, where random rotations in feature space support efficient quantization. A shared orthogonal right rotation, \((Q,K)\mapsto(QR,KR)\), preserves the score matrix \(S\).
In contrast, left rotation $Q \mapsto UQ$ acts on the token dimension and does not preserve $S$ in the deterministic way, but instead characterizes its statistical structure.

In high dimension this isotropic picture has a concrete implication for the entries
\(
S_{ij}={\langle q_i,k_j\rangle}/{\sqrt{d_{qk}}},
\)
where \(q_i\) and \(k_j\) are token-level query and key vectors.
When distinct-token directions are approximately independent and isotropic, their cosine similarities satisfy
\(
{\langle q_i,k_j\rangle}/{(\|q_i\|_2\|k_j\|_2)}\to 0 
\)
 for \(i\ne j\), 
in probability as the feature dimension grows.
This concentration describes angular geometry rather than the magnitude of the scaled scores, which also depends on query and key norms. It motivates studying score spectra separately from their orientation.
In our spectral surrogate model, we keep the singular values and replace the token-space orientation by the canonical representative \(\Sigma\).

\paragraph{Empirical observation.}
We support this hypothesis with empirical evidence from both vision and language models: a pretrained Vision Transformer evaluated on Imagenette \citep{Imagenette}, and GPT-2 evaluated on Wikitext-103.
Across these settings, the token-space singular matrices exhibit spectral behavior consistent with Haar-distributed orthogonal matrices.
In particular, their eigen-angle distributions are close to uniform on the unit circle, consistent with one spectral consequence of token-space isotropy.
Details are discussed in Section~\ref{experiment}.

\paragraph{From geometry to degeneration.}
Motivated by high-dimensional isotropy, we analyze
\(\operatorname{softmax}(\Sigma)\) and its causal counterpart,
where the diagonal surrogate \(\Sigma\) contains the singular values
of \(S\). We assume that bulk values lie below \(\log\ell\) and
spikes above it, with diverging additive margins.

In the bidirectional model, softmax normalization makes bulk rows
asymptotically uniform, while spike weights dominate the normalization
and concentrate rows on canonical basis vectors.
When the number of spikes satisfies \(k=o(\ell)\), the leading
\(k+1\) singular values and eigenvalues converge to \(1\), while the
rest vanish. We extend this spike--bulk analysis to causal attention
and generalized row-normalized attention maps.

\paragraph{Main contributions.}
Our contributions can be summarized as follows:
\begin{itemize}
\item We formulate a canonical diagonal score model motivated by token-space isotropy and provide empirical diagnostics of token-space geometry in pretrained transformers.
\item We characterize the eigenvalue and singular-value spectra of canonical bidirectional attention and show that, under spike--bulk separation, their empirical distributions concentrate on \(\{0,1\}\). Our results include a non-asymptotic characterization of the ordered eigenvalues and spectral profiles with explicit error bounds.
\item To the best of our knowledge, we provide the first joint characterization
    of eigenvalue and singular-value degeneration in \emph{causal} attention. Under the canonical spike--bulk score model, we derive exact eigenvalues
    and quantitative singular-value bounds, showing that bounded bulk
    scores yield harmonic bulk eigenvalue decay and at most logarithmically
    many bulk singular values above any fixed positive threshold.
\end{itemize}

\section{Related work}

\paragraph{Collapse in depth.}
\citet{attention_is_not_depth, anagnostidis2022signal} show that
repeated attention in pure attention networks progressively compresses
representations, producing doubly exponential rank collapse with depth. These results characterize the cumulative effect of composing attention
layers.

\paragraph{Collapse in width.}
In a different regime, \citet{mindthegap_width} analyze attention matrices in the width limit
by modeling the score matrix $S$ as an i.i.d.\ random matrix.
They show that the softmax attention matrix behaves like a random Markov matrix
and exhibits rank collapse driven by a spectral gap.
However, this analysis relies on the strong i.i.d.\ assumption on $S$,
which does not reflect the structured nature of trained transformers.

\citet{hayase2026gaussian} study the spectrum of attention matrices at initialization
through a Gaussian equivalence framework.
Unlike the i.i.d.\ model of \citet{mindthegap_width}, their analysis accounts for the
structured dependence in the score matrix~$S$, which arises even at initialization
from the shared input and weight matrices.
They provide a random matrix characterization of attention in the infinite-width limit,
capturing these correlations.
However, their results remain restricted to randomly initialized networks and do not address
the trained regime. 

Token-space isotropy observed in both Gaussian-initialized and trained
transformers motivates our canonical diagonal score surrogate, which
isolates a mechanism of degeneration under row-wise softmax normalization.
We derive asymptotic singular-value bounds and a non-asymptotic
eigenvalue characterization, and extend the analysis to causal attention
and generalized softmax attention with alternative activations.

\section{Main result}

\subsection{Setup}

Let $X \in \mathbb{R}^{\ell \times d}$ be the input sequence matrix,
where $\ell$ is the context length and $d$ is the hidden dimension.
Let $W_Q, W_K \in \mathbb{R}^{d \times d_{qk}}$ be the query and key projection matrices.
We define
\(
Q = X W_Q,  K = X W_K,
\)
and the (scaled) score matrix
\(
S =  QK^\top/\sqrt{d_{qk}} \in \mathbb{R}^{\ell \times \ell}.
\)
The attention matrix $A\in\mathbb{R}^{\ell \times \ell}$ is the row stochastic matrix given by the row-wise softmax:
\[
A_{ij} = \frac{\exp(S_{ij})}{\sum_{k=1}^\ell \exp(S_{ik})}.
\]

Token-space isotropy motivates examining the singular value decomposition
\(
S = U\Sigma V^\top,
\)
with \(U,V\in\mathrm{O}(\ell)\). We use \(\Sigma\) to define the following canonical attention model.

\begin{assumption}[Spectral canonical model]
\label{ass:diagonal-in-law}
Let \(s_1\ge s_2\ge \cdots \ge s_\ell\ge 0\) denote the singular values of the pre-softmax score matrix \(S\), and let
\(
\Sigma=\operatorname{diag}(s_1,\dots,s_\ell).
\)
The model attention matrices are defined by
\[
A_{\mathrm b}=\operatorname{softmax}(\Sigma),
\qquad A_{\mathrm c}=\operatorname{softmax}(\Sigma+M),
\]
where \(M\) is the causal mask defined below. We write \(A\) for the model under discussion. In the causal model, the singular values are placed in descending diagonal order.
\end{assumption}

\begin{assumption}[Spectral separation of \(\Sigma\)]
\label{ass:spectral-separation}
Let $s_1 \ge s_2 \ge \dots \ge s_\ell \ge 0$ be the diagonal entries of $\Sigma$. 
There exist an integer $k=k(\ell)\in\{0,\dots,\ell\}$ and
functions $f(\ell),g(\ell)\to\infty$, with $g(\ell)=o(\ell)$,
such that, for all sufficiently large $\ell$:
\begin{enumerate}
    \item \textbf{The spikes ($i\le k$):} The $k$ spike entries lie above the logarithmic normalization scale
    by a diverging margin, with $e^{s_i}\ge\ell f(\ell)$, and hence
    \[s_i=\Omega(\log(\ell f(\ell))).\] 

    \item \textbf{The bulk ($i>k$):} 
    The remaining $\ell - k$ bulk entries lie below the same scale
    by a diverging margin, with $e^{s_i}\le\ell/g(\ell)$, and hence
    \[s_i=O(\log(\ell/g(\ell))).\]
\end{enumerate}
The singular spectrum exhibits a strict gap between the bulk and the spikes, with no singular value $s_i$ lying between the two thresholds.
\end{assumption}

\begin{figure}[t]
    \centering
    \includegraphics[width=\linewidth]{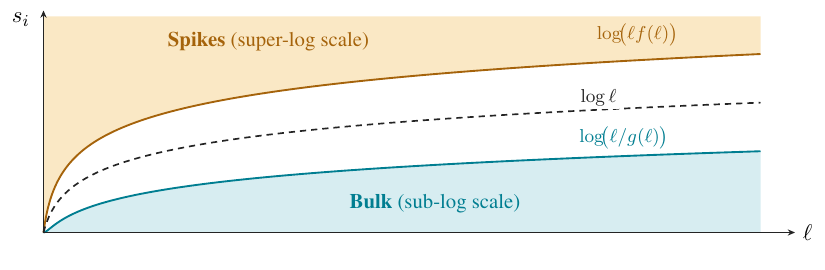}
    \caption{Qualitative spike--bulk separation under Assumption~\ref{ass:spectral-separation}.
    Spike and bulk scores satisfy \(s_i\ge\log(\ell f(\ell))\) and
    \(s_i\le\log(\ell/g(\ell))\), respectively; the open band between them contains no score entries.
    The dashed curve marks the normalization scale \(\log\ell\).}
    \label{fig:spike-bulk-regimes}
\end{figure}

The logarithmic threshold comes from the softmax normalization:
\(A_{ii}=e^{s_i}/(e^{s_i}+\ell-1)\). The diagonal weight \(e^{s_i}\)
competes with \(\ell-1\) unit off-diagonal weights, placing the transition
between averaging and diagonal concentration at \(s_i\approx\log\ell\)
(Figure~\ref{fig:spike-bulk-regimes}).

For a fixed head dimension \(d_{qk}\), the score matrix satisfies
\(k\le\operatorname{rank}(S)\le d_{qk}\).
Consequently, the number of spikes is sublinear in sequence length,
with \(k=o(\ell)\).

Motivated by isotropy, the canonical model retains the score singular
values and aligns the left and right singular directions with the
token axes, isolating the spectral effects of softmax normalization
and causal masking. Section~\ref{experiment} compares the model with
attention in pretrained transformers.


\subsection{Attention degeneration}
\label{subsec:bidirectional_attention}

In this subsection, \(A=A_{\mathrm b}=\operatorname{softmax}(\Sigma)\).
Throughout the remaining analysis, we use
\[
c_\ell=\frac{k}{\ell},\qquad
\varepsilon_\ell=\max\{f(\ell)^{-1},g(\ell)^{-1}\}
\]
for the spike fraction and the bidirectional approximation error scale, respectively.

\begin{theorem}[Attention Singular Degeneration]
\label{thm:singular-collapse}
Under Assumptions~\ref{ass:diagonal-in-law}
and~\ref{ass:spectral-separation}, as \(\ell\to\infty\), the following hold:
\begin{enumerate}
    \item \textbf{Singular-value profile.} Define
    \[
    \alpha_\pm(x)=\sqrt{1-\frac x2\pm\sqrt{x-\frac34x^2}},\qquad x\in[0,1].
    \]
    For \(1\le k<\ell\), the ordered deterministic profile is
    \[
    (\sigma_1^*,\ldots,\sigma_\ell^*)=
    \left(
    \alpha_+(c_\ell),
    \underbrace{1,\ldots,1}_{k-1},
    \alpha_-(c_\ell),
    \underbrace{0,\ldots,0}_{\ell-k-1}
    \right).
    \]
    The endpoint profiles are \((1,0,\ldots,0)\) for \(k=0\)
    and \((1,\ldots,1)\) for \(k=\ell\).
    The singular values of \(A\) satisfy the uniform approximation
    \[
    \max_{1\le i\le\ell}|\sigma_i(A)-\sigma_i^*|
    =O(\varepsilon_\ell).
    \]
    In particular, \(\sigma_i(A)=O(\varepsilon_\ell)\to0\) for \(i\ge k+2\).

    \item \textbf{Sublinear spikes.} If \(k=o(\ell)\), the leading
    \(k+1\) singular values converge uniformly to \(1\), with
    \[
    \max_{1\le i\le k+1}|\sigma_i(A)-1|
    =O\!\left(\varepsilon_\ell+\sqrt{c_\ell}\right)\to0.
    \]

    \item \textbf{Spectral degeneration.} If \(c_\ell\to c\in[0,1]\),
    the empirical singular-value distribution
    \(\mu_A^{\mathrm{sv}}:=\ell^{-1}\sum_{i=1}^\ell\delta_{\sigma_i(A)}\)
    converges weakly to \(c\delta_1+(1-c)\delta_0\).
\end{enumerate}
\end{theorem}

The proof of Theorem~\ref{thm:singular-collapse} gives the exact
diagonal-plus-rank-one decomposition
\begin{equation}
\label{eq:attention-decomposition}
A = D + z\mathbf{1}^\top,
\end{equation}
where $\mathbf{1}\in\mathbb{R}^{\ell}$ is the all-ones vector,
$Z_i=e^{s_i}+\ell-1$ is the row partition function,
$z=(z_1,\ldots,z_\ell)^\top$ with $z_i=Z_i^{-1}$, and
$D=\operatorname{diag}(d_1,\ldots,d_\ell)$ with $d_i=1-\ell z_i$.

When a bulk is present, all bulk rows perform the same averaging operation
in the limit and therefore contribute a single direction. Together with the \(k\) directions
retained by the spike rows, this shared averaging direction forms a
\((k+1)\)-dimensional subspace.

From decomposition~\eqref{eq:attention-decomposition}, we derive a closed-form characteristic equation and show that the canonical attention matrix has real eigenvalues despite its asymmetry.

\begin{theorem}[Attention Eigen Degeneration]
\label{thm:eigen-collapse}
Under Assumptions~\ref{ass:diagonal-in-law}
and~\ref{ass:spectral-separation}, write
\(d_1\ge\cdots\ge d_\ell\) for the diagonal entries of \(D\).
\begin{enumerate}
    \item \textbf{Exact spectral structure.} For every \(\ell\), the eigenvalues
    \(\lambda_1(A)\ge\cdots\ge\lambda_\ell(A)\) are real and lie in \([0,1]\),
    with \(\lambda_1(A)=1\). For \(2\le i\le\ell\), they satisfy
    \[
    \begin{cases}
    \lambda_i(A)=d_i, & \text{if }d_i=d_{i-1},\\
    d_i<\lambda_i(A)<d_{i-1}, & \text{if }d_i<d_{i-1}.
    \end{cases}
    \]
    For \(\ell>1\), this also gives the bounds
    \[
    \frac{\ell A_{ii}-1}{\ell-1}
    \le\lambda_i(A)\le
    \frac{\ell A_{i-1,i-1}-1}{\ell-1},\qquad 2\le i\le\ell.
    \]

    \item \textbf{Eigenvalue profile.} As \(\ell\to\infty\), the spike and bulk eigenvalues satisfy
    \[
    \begin{aligned}
    1-O(f(\ell)^{-1})
    &\le d_i\le\lambda_i(A)\le d_{i-1}<1,
    &&2\le i\le k,\\
    0&\le d_i\le\lambda_i(A)\le d_{i-1}
    =O(g(\ell)^{-1}),
    &&k+2\le i\le\ell.
    \end{aligned}
    \]
    If \(k<\ell\), the remaining eigenvalue satisfies
    \[
    \lambda_{k+1}(A)=1-c_\ell+O(\varepsilon_\ell),
    \]
    and hence converges to \(1-c\) whenever \(c_\ell\to c\in[0,1]\).
    In particular, if \(k=o(\ell)\), the leading \(k+1\) eigenvalues converge uniformly to \(1\).

    \item \textbf{Spectral degeneration.} If \(c_\ell\to c\in[0,1]\),
    the empirical eigenvalue distribution
    \(\mu_A^{\mathrm{eig}}:=\ell^{-1}\sum_{i=1}^\ell\delta_{\lambda_i(A)}\)
    converges weakly to \(c\delta_1+(1-c)\delta_0\).
\end{enumerate}
\end{theorem}

\begin{remark}[Bounded Bulk]
\label{rem:att-strong-assumption}
If the bulk scores satisfy \(s_i\le B\) for \(i>k\), with a constant \(B\ge0\) independent of \(\ell\), then
\[
0\le d_i=\frac{e^{s_i}-1}{e^{s_i}+\ell-1}
\le\frac{e^B-1}{\ell},\qquad i>k.
\]
Consequently, the \(O(g(\ell)^{-1})\) contributions in
Theorems~\ref{thm:singular-collapse}--\ref{thm:eigen-collapse} improve to
\(O(\ell^{-1})\), and the \(O(\varepsilon_\ell)\) profile errors become
\(O(\max\{f(\ell)^{-1},\ell^{-1}\})\).
\end{remark}

\subsection{Causal attention}
\label{subsec:causal_attention}

We now turn to causal attention, the masked variant used in decoder
architectures \citep{GPT1, GPT2,GPT3}.  The lower-triangular mask changes the
asymptotic operator structure: even benign bulk scores leave a triangular
averaging component rather than the same collapse seen in bidirectional
attention.
Throughout this subsection, \(A=A_{\mathrm c}=\operatorname{softmax}(\Sigma+M)\),
where the causal mask is
\[
    M_{i,j} = \begin{cases}
         -\infty, & \text{for } j > i\\
         0, & \text{else }, 
    \end{cases}
\]
Under Assumption~\ref{ass:spectral-separation}, the first \(k\) token positions
contain the spikes and the remaining \(\ell-k\) positions form the bulk.
Since the causal matrix is lower triangular, its eigenvalues are exactly its
diagonal entries.  The spike eigenvalues remain close to one, while harmonic
bulk decay follows once the relevant bulk scores are uniformly bounded.

\begin{theorem}[Eigenvalues of Causal Attention]
\label{thm:causal_eigenvalues}
Under Assumptions~\ref{ass:diagonal-in-law} and~\ref{ass:spectral-separation},
for any sequence length $\ell$, the eigenvalues of \(A\) are real and equal to its diagonal entries:
\[
\lambda_i(A) = \frac{\exp(s_i)}{\exp(s_i)+i-1}.
\]
The largest eigenvalue is exactly $\lambda_1 = 1$. As $\ell \to \infty$, 
\begin{itemize}
    \item The $k$ largest  eigenvalues converge to $1$, with speed \(O\!\left(\frac{1}{f(\ell)}\right) \).
    \item The \(r\) least eigenvalues decay harmonically:
    \[
    \lambda_i(A) = \Theta(1/i)
    \qquad \text{for} \ i \in \{\ell-r+1,\dots,\ell\},\] provided that, for some
    \(r\le \ell-k\), the last \(r\) bulk scores are uniformly bounded, i.e.,
    \(s_i=O(1)\) for \(i\in\{\ell-r+1,\dots,\ell\}\).  In particular, if all
    bulk scores \(s_i\), \(i>k\), are uniformly bounded, then \(r=\ell-k\).
    \item \textbf{Spectral degeneration.}
    If \(c_\ell\to c\in[0,1]\), the empirical eigenvalue distribution
    of \(A\) converges weakly to \(c\delta_1+(1-c)\delta_0\).
\end{itemize}
\end{theorem}

The singular values are subtler because triangularity no longer identifies
them. The next result describes their spectral degeneration under spectral
separation, with sharper bulk bounds when the bulk scores are uniformly
bounded.
\begin{theorem}[Singular Values of Causal Attention]
\label{thm:causal_singular_values}
Under Assumption~\ref{ass:diagonal-in-law}, \(A\) satisfies, for every sequence length \(\ell\),
\[
\frac1{3i}\le\sigma_i(A)\le\|A\|_{\mathrm{op}}\le\sqrt5,
\qquad 1\le i\le\ell.
\]
Under the additional spectral-separation assumption~\ref{ass:spectral-separation},
the following conclusions hold:
\begin{enumerate}
    \item \textbf{Spike subspace.}
    For \(k\ge1\) and all sufficiently large \(\ell\), there exists a
    $(k-1)$-dimensional subspace \(E_{\mathrm{sp}} \subset \mathbb{R}^\ell\) such that
    \[
    \left(1 - O\!\left(\frac{k}{\ell f(\ell)}\right)\right)\|v\|_2
    \le
    \|Av\|_2
    \le
    \left(1 + O\!\left(\frac{k}{\ell f(\ell)}\right)\right)\|v\|_2
    \qquad \text{for all } v\in E_{\mathrm{sp}}.
    \]
    Moreover, for \(k\ge1\),
    \[
    \sigma_k(A) \ge 1 - O\!\left(\frac{k}{\ell f(\ell)}\right).
    \]
    \item \textbf{Spectral degeneration.}
    If \(c_\ell\to c\in[0,1]\), the empirical singular-value
    distribution of \(A\) converges weakly to
    \(c\delta_1+(1-c)\delta_0\).
\end{enumerate}
If, in addition, there exists a constant \(B>0\), independent of \(\ell\),
such that \(s_i\le B\) for all \(i>k\), then there exists a constant
\(C_B>0\), depending only on \(B\), such that the following further
estimates hold for all sufficiently large \(\ell\):
\begin{enumerate}
    \setcounter{enumi}{2}
    \item \textbf{Bulk singular-value upper bounds.}
    For every ordered bulk index \(m \in \{1,\dots,\ell-k\}\),
    \[
    \sigma_{k+m}(A)
    \le
    C_{B}\min\!\left\{
    1,
    \frac{\ell-k-m+1}{k+m},
    \sqrt{\frac{\log(\ell+1)}{m}}
    \right\}.
    \]
    Moreover, for every fixed \(\varepsilon>0\),
    \(
    \#\{i>k:\sigma_i(A)>\varepsilon\}
    \le
    \frac{C_B\log(\ell+1)}{\varepsilon^2}.
    \)
    Consequently, for every sequence
    \(m_\ell\in\{1,\dots,\ell-k\}\) with \(m_\ell/\log\ell\to\infty\),
    \[
    \sigma_{k+m_\ell}(A)\to0.
    \]
\end{enumerate}
\end{theorem}

Weak spectral degeneration holds under spectral separation, while the
stronger logarithmic bounds use the additional bounded-bulk condition.
The causal mask gives the bulk a Hardy-type lower triangular averaging
structure, already visible when \(s_i\equiv0\).

\subsection{Encoder versus decoder}
Within the canonical model, bounded bulk scores give an \(O(1/\ell)\) bulk
term for bidirectional attention, whereas causal attention has at most
logarithmically many bulk singular values above any fixed threshold.
This contrast suggests a spectral
perspective on encoder and decoder representations: concentration in a few
directions may favor selective aggregation, while retaining additional
directions may support the range of continuations in sequential prediction.

\begin{figure}[tbp]
    \centering
    \includegraphics[width=\linewidth]{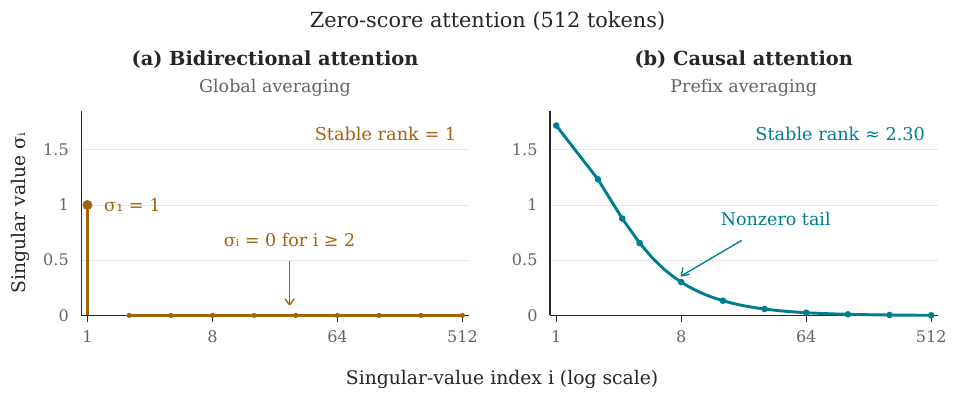}
    \caption{Ordered singular values of zero-score attention at \(\ell=512\).
    Bidirectional attention has rank one, whereas causal prefix averaging
    retains a nonzero singular-value tail.}
    \label{fig:zero-score-singular-spectra}
\end{figure}

The contrast is already visible when \(\Sigma=0\): bidirectional attention
returns the global average, while causal attention averages each prefix,
\[
A_{\mathrm b}=\frac{\mathbf1\mathbf1^\top}{\ell},
\qquad
(A_{\mathrm c})_{ij}=\frac{\mathbf1_{\{j\le i\}}}{i}.
\]
Here \(\|A_{\mathrm b}\|_{\mathrm F}^2=1\), whereas
\(\|A_{\mathrm c}\|_{\mathrm F}^2=\sum_{i=1}^{\ell}1/i\sim\log\ell\).
Their operator norms remain of constant order:
\(\|A_{\mathrm b}\|_{\mathrm{op}}=1\) and
\(1\le\|A_{\mathrm c}\|_{\mathrm{op}}\le2\).
Figure~\ref{fig:zero-score-singular-spectra} compares their ordered singular values.

To quantify this difference more precisely, we introduce the \emph{stable rank}
\(\srank(A)=\|A\|_{\mathrm F}^2/\|A\|_{\mathrm{op}}^2\)
and entropy-based \emph{effective ranks}. For a nonzero matrix \(A\) with
singular values \(\sigma_j(A)\), define
\[
\erank_p(A)=\exp\!\left(-\sum_j q_j^{(p)}\log q_j^{(p)}\right),
\qquad
q_j^{(p)}=\frac{\sigma_j(A)^p}{\sum_r\sigma_r(A)^p},
\quad p\in\{1,2\}.
\]
Here \(\erank_1\) is the effective rank of \citet{erank}, while
\(\erank_2\) uses squared singular values; we take \(0\log0=0\).
\begin{theorem}[Informal: Spectral ranks]
Consider canonical bidirectional and causal attention with a fixed number
\(k\) of spectrally separated spikes and bulk scores uniformly bounded by
\(B\). As \(\ell\to\infty\),
\[
\begin{aligned}
\srank(A_{\mathrm b}),\ \erank_2(A_{\mathrm b})&=k+1+o(1),\\
\srank(A_{\mathrm c}),\ \erank_2(A_{\mathrm c})&=\Theta_B(\log\ell).
\end{aligned}
\]
Moreover, \(\erank_1(A_{\mathrm c})=\Omega_B((\log\ell)^3)\), while
\(\erank_1(A_{\mathrm b})=k+1+o(1)\) when its linear bulk mass and
entropy moment vanish. 
\end{theorem}

Precise conditions and bounds are given
in Theorems~\ref{thm:attention-rank-proxies} and~\ref{thm:causal-rank-proxies}
in Appendix~\ref{app:rank-proxies}.
These bounds distinguish the two attention patterns despite their common
limiting empirical spectral law \(\delta_0\). Relative to context length,
the stable ranks decay as \(1/\ell\) and \(\log\ell/\ell\), respectively,
quantifying how causal masking slows spectral degeneration as context length
grows.
Further technical remarks on the causal singular-value bounds are collected in Appendix~\ref{app:causal-singular-remarks}. 

\section{Experiment}
\label{experiment}

Our experiments on pretrained GPT-2 and ViT connect the spectral model to
trained attention through three questions: how token-space singular factors
are oriented, how score strength is distributed across singular directions,
and how real and canonical attention spectra compare after softmax. 

\subsection{Eigen-angle diagnostics of token-space singular factors}

\paragraph{Imagenette data}
We examine the pretrained ViT-B/16 model with SWAG weights
\citep{SWAG_2022_CVPR}, using the validation
split of Imagenette.  Each $384 \times 384$ image is mapped to a token matrix
$X \in \mathbb{R}^{577 \times 768}$ at each layer.  For every layer input, we
compute a full singular value decomposition
\(
    X = U_X \Sigma_X V_X^\top,
\)
and examine the square token-space singular factor $U_X$. Its eigenvalues lie
on the unit circle, so we write $\lambda_j(U_X)=e^{i\theta_j}$ with
$\theta_j\in[-\pi,\pi]$. Haar orthogonal matrices have asymptotically uniform
empirical eigen-angles \cite[Theorem~3]{Diaconis94}, giving a necessary
asymptotic diagnostic for Haar behavior. We measure discrepancies from the
uniform distribution using Kullback-Leibler divergence, Kolmogorov distance,
and angular Wasserstein distance.
Figure~\ref{fig:vit-layer-distance-boxplot} 
shows that the measured
discrepancies are consistently close to zero across ViT layers, supporting the
isotropy motivation for the spectral canonical model in the vision setting.

\paragraph{Wikitext data}
We apply the same diagnostic to the token-space singular factor $U_X$ of
GPT-2 hidden-state inputs on the Wikitext-103 validation split
\citep{merity2017pointer}, with sequence lengths from $17$ to $471$ tokens.
\begin{figure}[t]
  \centering
  \includegraphics[width=\linewidth]{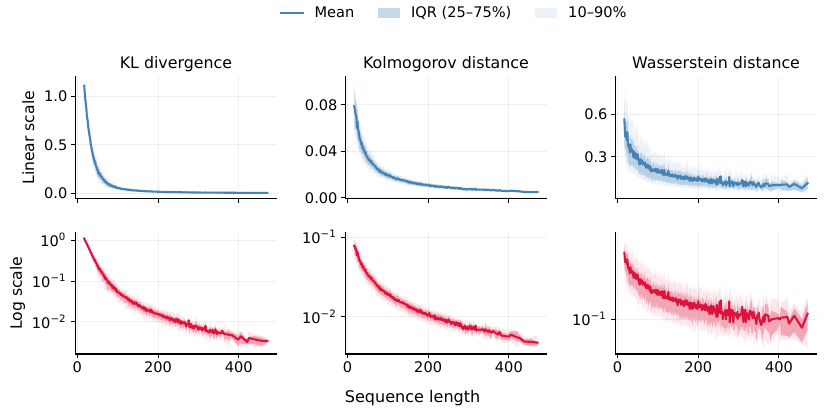}
\caption{Discrepancies between the GPT-2 token-space singular eigen-angle distribution and the uniform distribution on $[-\pi,\pi]$ as a function of sequence length on the Wikitext-103 validation split.  Linear- and log-scale summaries are shown for Kullback-Leibler divergence, Kolmogorov distance, and Wasserstein distance, with means, interquartile ranges, and 10--90\% bands. Kullback-Leibler divergence is calculated by histogram binning with $50$ bins.}
  \label{fig:wikitext-distance-metrics}
\end{figure}
Figure~\ref{fig:wikitext-distance-metrics} shows decreasing discrepancies
with sequence length, indicating increasingly uniform empirical
eigen-angle distributions over the tested range.

\subsection{Effective rank of the score matrix}

\begin{figure}[htbp]
 \centering
 \includegraphics[width=\linewidth]{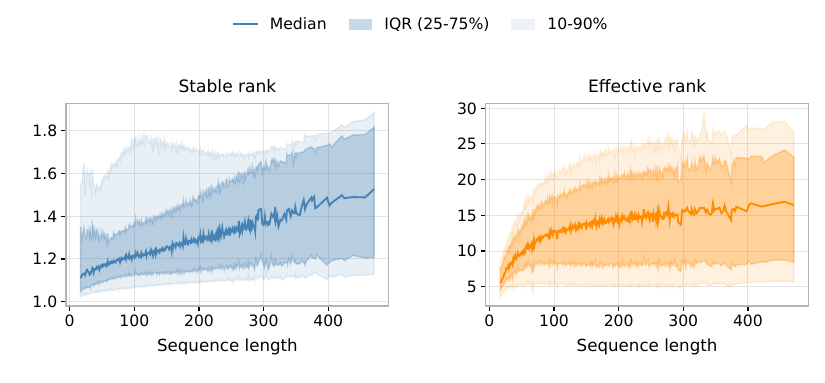}
 \caption{Stable rank and effective rank of GPT-2 score matrices on the
 Wikitext-103 validation split as a function of sequence length. Curves show
 the median trend, with interquartile ranges and wider percentile bands.}
 \label{fig:score-rank-wikitext}
\end{figure}

Our theory allows a nonzero bulk, but the spike--bulk analysis is most
informative when a few dominant singular directions capture most of the score
matrix energy.  We therefore measure the intrinsic rank of GPT-2 score matrices
on the Wikitext-103 validation split as $\ell$ varies.  
Figure~\ref{fig:score-rank-wikitext} 
shows that both rank surrogates remain
moderate across the tested sequence lengths rather than scaling linearly with
$\ell$, motivating the spike--bulk structure in Assumption~\ref{ass:spectral-separation}.

\subsection{Real-versus-canonical attention spectra}
\label{subsec:real-canonical-comparison}

We examine the spectral canonical model by directly comparing real
attention with the attention induced by the score singular values. For each
head, we construct
\(A_{\mathrm{real},h}=\operatorname{softmax}_{\mathrm{row}}(S_h+M)\) and
\(A_{\Sigma,h}=\operatorname{softmax}_{\mathrm{row}}(\Sigma_h+M)\), using the
same mask and placing the score singular values in descending diagonal
order. Here $M$ is the causal mask for GPT-2 and zero for ViT.
These score and attention diagnostics are computed separately for each
head of the intact pretrained networks, with aggregation only after the
spectral measurements are obtained (see Appendix~\ref{app:head-heterogeneity}
for head-resolved summaries).
Measure the Wasserstein distance between their empirical
singular-value distributions:
\[
W_1(\mu_{A_{\mathrm{real},h}}^{\mathrm{sv}},\mu_{A_{\Sigma,h}}^{\mathrm{sv}})
=\frac{1}{\ell}\sum_{j=1}^{\ell}
\left|\sigma_j(A_{\mathrm{real},h})-\sigma_j(A_{\Sigma,h})\right|.
\]

\begin{figure}[htbp]
 \centering
 \includegraphics[width=\linewidth]{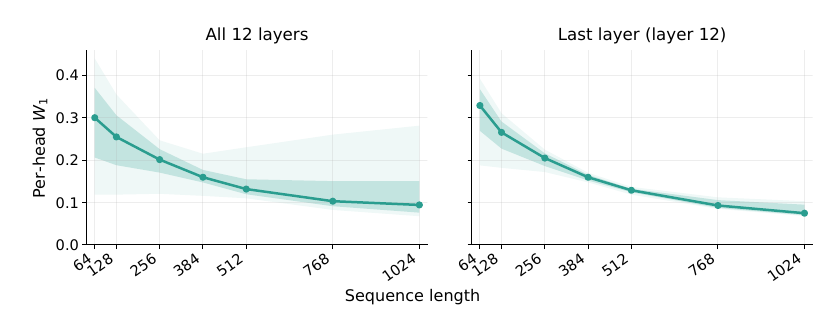}
 \caption{GPT-2 real-versus-canonical attention spectral discrepancy across
 sequence lengths, pooled over all layers (left) and restricted to the last
 layer (right). Each length uses nested prefixes of the same 12 documents,
 with 12 heads per layer. Curves show median $W_1$; IQR and 10--90\% bands
 summarize variation across the paired observations.}
 \label{fig:gpt2-real-canonical-w1}
\end{figure}

\begin{figure}[htbp]
 \centering
 \includegraphics[width=0.7\linewidth]{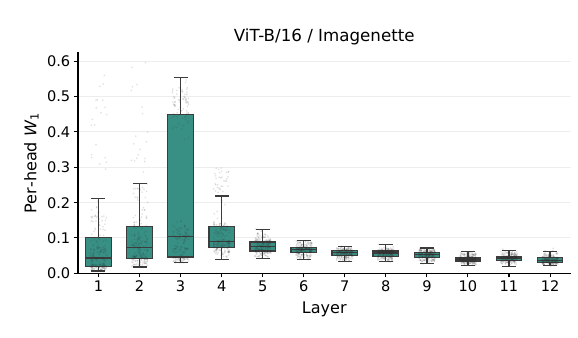}
 \caption{ViT real-versus-canonical attention spectral discrepancy across
 layers at sequence length $577$, with 20 images and 12 heads per layer.
 Boxes show medians and IQRs, with whiskers
 extending to $1.5$ times the IQR. All layer medians are below $0.104$.}
 \label{fig:vit-real-canonical-w1}
\end{figure}

Figure~\ref{fig:gpt2-real-canonical-w1} shows a substantial decrease in
spectral discrepancy as GPT-2 context length grows from 64 to 1,024.
The median distance falls from $0.3002$ to $0.0942$ across all layers and
from $0.3290$ to $0.0748$ in the last layer. At $\ell = 577$, ViT
has layerwise medians between $0.0367$ and $0.1037$, with greater variation
across image--head pairs in early layers
(Figure~\ref{fig:vit-real-canonical-w1}). These observations support bulk
spectral similarity, with decreasing discrepancy over the tested GPT-2 lengths.

Rank proxies complement this distributional comparison by measuring
spectral concentration. At GPT-2 length 1,024, the pooled stable-rank
medians are $1.18$ for real attention and $27.51$ for canonical attention;
the corresponding $\erank_1$ medians are $314.06$ and $95.12$.
Thus proximity of equally weighted spectral distributions coexists with
different concentration and entropy profiles.
The sampling and numerical
details and the relation to leading spectral modes are discussed in Appendix~\ref{app:real-canonical-comparison}.

\subsection{Score-level geometry and spectral structure}
\label{subsec:score-geometry}

We apply the eigen-angle diagnostics directly to the actual scores,
decomposing \(S_h=U_{S_h}\Sigma_hV_{S_h}^{\top}\) independently for every
sample, layer, and head. The analysis covers 1,646 GPT-2 sequences and
3,925 ViT images, with all 12 layers and 12 heads. For both GPT-2 factors,
the mean angular $W_1$ decreases from approximately $0.31$ in the
$17$--$64$ length band to $0.10$ in the $385$--$471$ band; the pooled
ViT mean is approximately $0.08$ for each factor. KL and Kolmogorov
discrepancies show similar GPT-2 length trends and remain small across ViT
layers (Figures~\ref{fig:score-gpt2-angles} and~\ref{fig:score-vit-angles}).
These measurements
extend the near-uniform eigen-angle patterns from hidden-state inputs to
the computed score singular factors, supporting the isotropy motivation
for Assumption~\ref{ass:diagonal-in-law}.

Figures~\ref{fig:gpt2-score-spectrum} and~\ref{fig:vit-score-spectrum}
complement these geometric diagnostics with logarithmic score spectra for
one input per model. Each layer pools nonzero singular values after
independent per-head decomposition; an empirical-gap split highlights a
high-value group above a lower bulk. Together with the modest score-rank
proxies in Figure~\ref{fig:score-rank-wikitext}, these descriptive spectra
motivate a model with a few dominant score directions. The quantitative
thresholds in Assumption~\ref{ass:spectral-separation} specify the asymptotic
separation regime analyzed by the theory. The structural bound
\(k\le\operatorname{rank}(S_h)\le d_{qk}=64\) supplies a length-independent
upper bound on the number of spikes which depends on the
input and learned projections.

\section{Conclusion}

Despite the empirical success of self-attention, a spectral analysis of
attention matrices under consistent assumptions from initialization
to the trained regime has been largely missing.  Our central observation is that
the token-space singular factors exhibit nearly uniform empirical eigen-angle
distributions across a range of Transformer-based models, including both encoder and decoder
architectures, in long-sequence regimes.  This empirical property
motivates our spectral canonical model for the score matrix, while the
observed effective ranks motivate the spike--bulk assumption on its singular values.
Together these assumptions make the spectrum of the attention matrix amenable to
non-asymptotic and asymptotic analysis
that
characterizes degeneration, or rank collapse, within the canonical model.
The theory 
provides a model mechanism for spectral compression observed in practice, offering one perspective
on why this architecture has been so effective in LLMs.

Future work could study how alignment of dominant token subspaces
across heads and the value and output projections shape the
effective dimension of multi-head attention, including the effects
of residual connections and normalization.
Isotropy and spike--bulk diagnostics could also test for analogous
spectral compression in the input-dependent recurrences of selective
state-space models such as Mamba \citep{gu2023mamba}. This would support
comparisons with attention at the level of their learned sequence
operators.



\begin{ack}
We thank the reviewers for their constructive remarks.
Jianfeng Yao's research is supported by NSFC
Grant RFIS-A10120230617, the Guangdong Pearl River Talent Program Grant
2023JC10X022, and the Guangdong Provincial Key Laboratory of Mathematical
Foundations for Artificial Intelligence (2023B1212010001). The authors
declare no competing interests.
\end{ack}
\bibliographystyle{plainnat}
\bibliography{ref}

\appendix

\section{Scope of the canonical reduction}
\label{app:model-scope}

The diagonal surrogate is an additional modeling choice. Neither score
singular values nor Haar-distributed singular-vector orientations alone
justify replacing the attention matrix by its canonical counterpart.
The distinction matters both before and after applying a causal mask.

\paragraph{Equality in law and orientation averaging.}
The statement \(S\overset d=\Sigma\), with \(\Sigma\) diagonal and
nonnegative, would force \(S\) to be diagonal and nonnegative almost
surely. It cannot express a nontrivial bi-orthogonally invariant score
law. In particular, if \(S=U\Sigma V^\top\), where \(U,V\) are independent
Haar orthogonal matrices independent of \(\Sigma\), the average over
orientations is not \(\Sigma\). When integrable,
\(\mathbb E[S\mid\Sigma]=0\) by sign symmetry. For both attention patterns, the conditional means are
\[
\mathbb E[\operatorname{softmax}(S)\mid\Sigma]
=\ell^{-1}\mathbf1\mathbf1^\top,
\qquad
\mathbb E[\operatorname{softmax}(S+M)\mid\Sigma]_{ij}
=\frac{\mathbf1_{\{j\le i\}}}{i}.
\]
Indeed, conditional invariance under column permutations makes the
expected weights equal within each row's unmasked positions, and each
row sums to one. For a causal row, use permutations of its first \(i\)
columns. These identities require only conditional column-permutation
invariance, and concern mean attention matrices, not expected singular
values or effective ranks.

\paragraph{A score-spectrum counterexample.}
For \(\ell\ge2\), let
\[
\Sigma=t_\ell e_1e_1^\top,
\qquad S=\frac{t_\ell}{\ell}\mathbf1\mathbf1^\top,
\qquad t_\ell=2\log\ell.
\]
Both score matrices have singular values \((t_\ell,0,\ldots,0)\),
satisfying Assumption~\ref{ass:spectral-separation} with one spike.
Every row of \(S\) is constant, so
\(\operatorname{softmax}(S)=\ell^{-1}\mathbf1\mathbf1^\top\) has rank one.
In contrast, Theorem~\ref{thm:singular-collapse} gives two singular
values tending to one for \(\operatorname{softmax}(\Sigma)\).
Thus score singular values alone do not determine even the limiting
number of nonvanishing attention singular values. This example
illustrates orientation dependence; it does not assume Haar orientations.

\paragraph{The causal ordering is part of the model.}
For \(\ell=2\), causal attention with diagonal scores \((t,0)\) has
eigenvalues \(1,1/2\), whereas scores \((0,t)\) give eigenvalues
\(1,e^t/(e^t+1)\). The score singular values are identical.
Consequently, placing the descending singular values along the diagonal
and then applying a fixed causal mask also fixes their token positions.
The causal results are conditional on that convention.

\paragraph{Interpretation of the diagnostics.}
Nearly uniform eigen-angles of a chosen SVD factor are consistent with a
necessary asymptotic consequence of a Haar law; they do not establish
the joint Haar law, independence of left and right factors, or independence
from singular values. SVD signs and nullspace completions also affect
this diagnostic. Similarly, small score rank proxies do not establish
the explicit exponential thresholds of
Assumption~\ref{ass:spectral-separation}.
Appendix~\ref{app:empirical-scope} records these empirical qualifications.
For example, the small Wasserstein distance between empirical spectral
laws does not control their leading singular values or rank proxies.
A theorem transferring the canonical results to general score matrices
would need additional control of the orientation-dependent softmax
error; such a transfer is not proved here.

\paragraph{From individual heads to multi-head representations.}
For $n_h$ heads, let $A_h\in\mathbb R^{\ell\times\ell}$ be the attention
matrix and $V_h\in\mathbb R^{\ell\times d_v}$ the value matrix of head
$h\in\{1,\ldots,n_h\}$. Writing $Z_h=A_hV_h$ and $Z=[Z_1|\cdots|Z_{n_h}]$, the multi-head
output before adding any output bias is
\[
Y=[A_1V_1|\cdots|A_{n_h}V_{n_h}]W_O=ZW_O,
\qquad W_O\in\mathbb R^{n_h d_v\times d_{\mathrm{out}}}.
\]
The token-space Gram matrices satisfy
\[
ZZ^\top=\sum_{h=1}^{n_h} A_hV_hV_h^\top A_h^\top,
\qquad
YY^\top=ZW_OW_O^\top Z^\top.
\]
These identities connect the headwise spectral picture to the combined
representation. Before the output projection, the column space is the
span of the head-output column spaces: aligned subspaces can retain a
common low-dimensional representation, whereas distinct subspaces can
increase the dimension of the combined span. The singular values also
depend on the value matrices and the weighting induced by $W_O$.
Consequently, headwise attention spectra alone do not determine the
singular spectrum of $Y$. The head-resolved measurements in
Appendix~\ref{app:head-heterogeneity} describe variation in spectral
concentration and canonical discrepancy; a theory of the combined output
would additionally require a joint description of the head-output
subspaces and projections. Extending this description through residual
connections and normalization provides a further question at the
Transformer-block level.

\section{Generalized softmax attention}
Recent work has explored alternative attention activations to reduce
computational cost. The canonical results extend to positive activations
followed by row normalization. We leave unnormalized attention mechanisms
to complementary analyses such as \citep{liao2026incontext}.

Let \(\gamma:\mathbb R\to(0,\infty)\) be nondecreasing on
\([0,\infty)\). Multiplying \(\gamma\) by a positive constant does not
change the normalized attention, so take \(\gamma(0)=1\). Define
\[
A^\gamma_{ij}
=\frac{\gamma(S_{ij})}{\sum_{r=1}^{\ell}\gamma(S_{ir})}.
\]
For causal attention, masked entries are set to zero and only unmasked
entries enter the row normalization.

Under Assumption~\ref{ass:diagonal-in-law}, evaluate this map on
\(\Sigma\). Set
\[
\widetilde s_i=\log\gamma(s_i),\qquad
\widetilde\Sigma=\operatorname{diag}(\widetilde s_1,\ldots,
\widetilde s_\ell).
\]
Then \(\widetilde s_1\ge\cdots\ge\widetilde s_\ell\ge0\), and the
canonical generalized attention matrices are exactly
\[
A^\gamma_{\mathrm b}=\operatorname{softmax}(\widetilde\Sigma),
\qquad
A^\gamma_{\mathrm c}=\operatorname{softmax}(\widetilde\Sigma+M).
\]
Thus every canonical spectral and rank-proxy result applies with
\(s_i\) replaced by \(\widetilde s_i\). In particular, spectral
separation becomes
\[
\gamma(s_i)\ge\ell f(\ell)\quad(i\le k),\qquad
\gamma(s_i)\le\ell/g(\ell)\quad(i>k),
\]
with the same conditions on \(f\) and \(g\). Bounded-bulk conclusions
require uniformly bounded \(\gamma(s_i)\) in the bulk; \(s_i\le B\)
implies this through \(\gamma(s_i)\le\gamma(B)\).
The transition occurs when \(\gamma(s_i)\) is of order \(\ell\).
Neither strict monotonicity nor continuity nor an inverse activation is
needed. Bounded activations cannot meet the spike threshold for any
nonempty spike set, but still admit the zero-spike conclusions.

\section{Proofs of main results}

\subsection{Proof of Theorem~\ref{thm:singular-collapse}}
\label{app:proof-thm:singular-collapse}
\begin{proof}
\textbf{Singular-value profile.}
Set \(z_i=(e^{s_i}+\ell-1)^{-1}\), \(d_i=1-\ell z_i\),
\(D=\operatorname{diag}(d_i)\),
\(J=\ell^{-1}\mathbf1\mathbf1^\top\), and \(P=I-J\).
Direct row normalization gives the exact identities
\[
A=D+z\mathbf1^\top=J+DP,
\qquad d_i=\frac{e^{s_i}-1}{e^{s_i}+\ell-1}\in[0,1).
\]
Let \(D^\star=\operatorname{diag}(\mathbf1_k,\mathbf0_{\ell-k})\)
and \(A^\star=J+D^\star P\). Spectral separation implies
\[
1-d_i\le f(\ell)^{-1}\quad(i\le k),\qquad
d_i\le g(\ell)^{-1}\quad(i>k).
\]
Since \(P\) is an orthogonal projection,
\begin{equation}
\label{eq:canonical-operator-perturbation}
\|A-A^\star\|_{\mathrm{op}}
=\|(D-D^\star)P\|_{\mathrm{op}}
\le\max\{f(\ell)^{-1},g(\ell)^{-1}\}=\varepsilon_\ell.
\end{equation}
It remains only to compute the singular values of \(A^\star\).

For \(1\le k<\ell\), its first \(k\) rows are coordinate vectors
and its remaining rows equal \(\ell^{-1}\mathbf1^\top\). Thus
\[
A^\star(A^\star)^\top=
\begin{pmatrix}
I_k&\ell^{-1}\mathbf1_{k\times(\ell-k)}\\
\ell^{-1}\mathbf1_{(\ell-k)\times k}
&\ell^{-1}\mathbf1_{(\ell-k)\times(\ell-k)}
\end{pmatrix}.
\]
The zero-sum spike subspace has eigenvalue one with multiplicity \(k-1\),
and the zero-sum bulk subspace has eigenvalue zero with multiplicity
\(\ell-k-1\). On the remaining two-dimensional subspace, use the
orthonormal basis
\(k^{-1/2}(\mathbf1_k,0)\),
\((\ell-k)^{-1/2}(0,\mathbf1_{\ell-k})\).
The corresponding symmetric matrix is
\[
\begin{pmatrix}1&\sqrt{c_\ell(1-c_\ell)}\\
\sqrt{c_\ell(1-c_\ell)}&1-c_\ell\end{pmatrix},
\]
whose eigenvalues are
\[
a_\pm(c_\ell)=1-\frac{c_\ell}{2}\pm\sqrt{c_\ell-\frac34c_\ell^2}=\alpha_\pm(c_\ell)^2.
\]
For \(0<c_\ell<1\), \(a_+(c_\ell)>1>a_-(c_\ell)>0\), so the two
exceptional singular values occupy positions \(1\) and \(k+1\), giving
the ordered profile in the theorem. The endpoint profiles follow from
\(A^\star=J\) when \(k=0\) and \(A^\star=I\) when \(k=\ell\).
Weyl's inequality and~\eqref{eq:canonical-operator-perturbation} prove
the uniform approximation, including the vanishing tail.

\textbf{Sublinear spikes.}
If \(k=o(\ell)\), then \(a_\pm(c_\ell)=1+O(\sqrt{c_\ell})\), so
\(\alpha_\pm(c_\ell)=1+O(\sqrt{c_\ell})\). The other leading profile values
are exactly one. Combining this with the uniform approximation gives
the stated bound for all \(1\le i\le k+1\). For \(k=0\), the same
conclusion follows from \(\sigma_1^*=1\).

\textbf{Spectral degeneration.}
The singular profile has
\(k+O(1)\) unit values, \(\ell-k+O(1)\) zeros, and at most two
exceptional values, all uniformly bounded. Thus, if \(c_\ell\to c\in[0,1]\),
its empirical distribution converges weakly to
\(c\delta_1+(1-c)\delta_0\), including the endpoint profiles
\(k=0,\ell\). The uniform singular-value approximation transfers this
weak limit to \(\mu_A^{\mathrm{sv}}\).
\end{proof}

\subsection{Proof of Theorem~\ref{thm:eigen-collapse}}
\label{app:proof-thm:eigen-collapse}
\begin{proof}
\textbf{Exact spectral structure.}
Retain \(D,J,P,z\) from the preceding proof. The case \(\ell=1\) is
immediate, so assume \(\ell>1\). Let
\(u=\ell^{-1/2}\mathbf1\) and choose an orthonormal basis matrix
\(Q\in\mathbb R^{\ell\times(\ell-1)}\) for \(u^\perp\).
In the orthonormal basis \([u,Q]\),
\[
[u,Q]^\top A[u,Q]
=\begin{pmatrix}1&u^\top DQ\\0&Q^\top DQ\end{pmatrix}.
\]
Consequently the eigenvalues of \(A\) are one together with those of
the symmetric compression \(Q^\top DQ\). Since \(0\le D<I\), they
are real and belong to \([0,1]\), with a simple largest eigenvalue
\(1\). Cauchy interlacing for a codimension-one compression gives
\[
d_i\le\lambda_i(A)\le d_{i-1},\qquad 2\le i\le\ell.
\]
The relation \(A_{ii}=\ell^{-1}+(1-\ell^{-1})d_i\) yields the
displayed bounds in terms of \(A\)'s diagonal.

For the strict alternative and characteristic equation, applying the
determinant lemma to~\eqref{eq:attention-decomposition} gives, away from the poles \(d_i\),
\begin{equation}
\label{eq:determinant_lemma}
\det(\lambda I-A)
=\prod_{i=1}^{\ell}(\lambda-d_i)
\left(1-\sum_{i=1}^{\ell}\frac{z_i}{\lambda-d_i}\right).
\end{equation}
Group equal diagonal values into distinct poles \(\delta_r\), with
multiplicities \(m_r\) and residues
\(w_r=\sum_{i:d_i=\delta_r}z_i>0\).  The secular factor is then
\(1-\sum_r w_r/(\lambda-\delta_r)\).  Its nonzero residue at each
pole cancels exactly one of the \(m_r\) polynomial factors, so
\(\delta_r\) has algebraic multiplicity exactly \(m_r-1\).
The secular function is strictly increasing between poles and runs from
\(-\infty\) to \(+\infty\) between adjacent distinct poles.  It
has one further root above the largest pole and none below the smallest.
This accounts for all \(\ell\) roots and proves the stated strict
interlacing alternative: equality occurs precisely when \(d_i=d_{i-1}\);
otherwise \(\lambda_i(A)\) lies strictly between these two entries.

\textbf{Eigenvalue profile.}
Compare with \(D^\star\) above.
The associated ideal matrix \(A^\star\) is block lower triangular,
with diagonal blocks \(I_k\) and
\(\ell^{-1}\mathbf1_{(\ell-k)\times(\ell-k)}\).
For \(k<\ell\), its ordered eigenvalue list is
\[
\underbrace{1,\ldots,1}_{k\text{ times}},\quad
1-c_\ell,\quad
\underbrace{0,\ldots,0}_{\ell-k-1\text{ times}}.
\]
For \(k=\ell\), it consists of \(\ell\) ones.
Both non-Perron spectra are eigenvalues of symmetric compressions, and
\[
\|Q^\top(D-D^\star)Q\|_{\mathrm{op}}
\le\max\{f(\ell)^{-1},g(\ell)^{-1}\}=\varepsilon_\ell.
\]
Hermitian Weyl's inequality therefore proves the intermediate-eigenvalue
estimate uniformly for \(0\le k<\ell\). For \(k=0\) this is simply
\(\lambda_1=1\). The sharper spike error \(O(f(\ell)^{-1})\)
and bulk error \(O(g(\ell)^{-1})\) follow directly from interlacing
and the bounds on \(d_i\) in the preceding proof. If \(k=o(\ell)\),
then \(c_\ell\to0\), and the spike and intermediate estimates together
give uniform convergence of the leading \(k+1\) eigenvalues to one.

\textbf{Spectral degeneration.}
For \(k<\ell\), the ideal eigenvalue profile differs from \(k\) ones
and \(\ell-k\) zeros in only one entry, of empirical mass \(1/\ell\);
for \(k=\ell\), the profiles coincide. Hence, if
\(c_\ell\to c\in[0,1]\), its empirical distribution converges weakly
to \(c\delta_1+(1-c)\delta_0\). The same Hermitian perturbation
bound gives uniform approximation of the ordered eigenvalues and
therefore the asserted weak limit for \(\mu_A^{\mathrm{eig}}\).
\end{proof}

\begin{remark}[A sharper intermediate-eigenvalue bound]
For any nonnegative diagonal scores and \(1\le k<\ell\),
\[
(1-c_\ell)d_k+c_\ell d_\ell
\le\lambda_{k+1}(A)
\le(1-c_\ell)d_1+c_\ell d_{k+1}.
\]
Indeed, sandwich \(D\) between the two-block diagonal matrices with
block values \((d_k,d_\ell)\) and \((d_1,d_{k+1})\).
The \(k\)-th largest eigenvalue of the compression of
\(\operatorname{diag}(aI_k,bI_{\ell-k})\), for \(a\ge b\), to
\(\mathbf1^\perp\) is \((1-c_\ell)a+c_\ell b\); monotonicity of eigenvalues
under the Loewner order proves the bounds.  Under spectral separation,
they give the sharper one-sided estimate
\[
-\frac{1-c_\ell}{f(\ell)}
\le\lambda_{k+1}(A)-(1-c_\ell)
\le\frac{c_\ell}{g(\ell)}.
\]
In particular, if \(k=o(\ell)\), the spike condition alone implies
that the leading \(k+1\) eigenvalues converge uniformly to one:
the lower bound is at least \((1-c_\ell)d_k\to1\), while all eigenvalues
are at most one.  No bulk bound is needed for this head convergence;
when \(k=0\), it is simply \(\lambda_1=1\).
\end{remark}

\subsection{Proof of Theorem~\ref{thm:causal_eigenvalues}}
\label{app:proof-thm:causal_eigenvalues}
\begin{proof}
Write \(a_i=e^{s_i}\). The canonical causal matrix is lower triangular,
so its eigenvalues are its diagonal entries:
\[
\lambda_i(A)=\frac{a_i}{a_i+i-1}
=1-\frac{i-1}{a_i+i-1}.
\]
These are already in decreasing order: \(a_i\) is nonincreasing, so
\((i-1)/a_i\) is nondecreasing. In particular, \(\lambda_1=1\).
For \(i\le k\),
\[
0\le1-\lambda_i(A)\le\frac{i-1}{\ell f(\ell)}
\le\frac{k}{\ell f(\ell)},
\]
which also sharpens the stated \(O(1/f(\ell))\) spike error.

For spectral degeneration, put \(t_\ell=\ell/g(\ell)\ge1\) for all
sufficiently large \(\ell\). Every bulk row satisfies
\(\lambda_i(A)\le\min\{1,t_\ell/i\}\), since \(a_i+i-1\ge i\).
Splitting the sum at \(\lceil t_\ell\rceil\) gives
\(\sum_{i=1}^\ell\min\{1,t_\ell^2/i^2\}\le3t_\ell\).
Together with the spike estimate, this yields
\[
\frac1\ell\sum_{i=1}^\ell
\bigl(\lambda_i(A)-\mathbf1_{\{i\le k\}}\bigr)^2
\le\frac{k^3}{\ell^3f(\ell)^2}+\frac3{g(\ell)}=o(1).
\]
The empirical eigenvalue distribution therefore approaches
\(c_\ell\delta_1+(1-c_\ell)\delta_0\) in mean-square transport distance.
If \(c_\ell\to c\in[0,1]\), it therefore converges weakly to
\(c\delta_1+(1-c)\delta_0\), without any bounded-bulk assumption.

Finally, for every bounded-score bulk row, \(1\le a_i\le C\) gives
\(1/i\le\lambda_i(A)\le C/i\), proving the harmonic bounds.
This bounded-score condition cannot be dropped: if \(k=0\) and
\(a_i=\sqrt\ell\) for every \(i\), spectral separation holds with
\(g(\ell)=\sqrt\ell\), but
\(\lambda_\ell(A)=\sqrt\ell/(\sqrt\ell+\ell-1)\asymp\ell^{-1/2}\).
Thus the empirical eigenvalue distribution still converges to \(\delta_0\),
while uniform harmonic decay fails.
\end{proof}

\subsection{Proof of Theorem~\ref{thm:causal_singular_values}}
\label{app:proof-thm:causal_singular_values}
\begin{proof}
Write \(a_i=e^{s_i}\ge1\) and \(z_i=(a_i+i-1)^{-1}\). Then
\[
A_{ij}=\begin{cases}
z_i,&j<i,\\
a_i z_i,&j=i,\\
0,&j>i.
\end{cases}
\]
\paragraph{Step 1: universal singular-value bounds.}
Let \(\mathsf H_\ell\) be the Ces\`aro matrix, \((\mathsf H_\ell)_{ij}=i^{-1}\mathbf1_{\{j\le i\}}\),
and define \(\theta_i=(a_i-1)/(a_i+i-1)\in[0,1]\). We have the exact identity
\[
A=\operatorname{diag}(\theta_i)
+\bigl(I-\operatorname{diag}(\theta_i)\bigr)\mathsf H_\ell.
\]
By convexity of the square and the Hardy inequality
(Lemma~\ref{lem:hardy_l2}),
\[
\|Ax\|_2^2
\le\sum_i\bigl(\theta_i|x_i|^2+(1-\theta_i)|(\mathsf H_\ell x)_i|^2\bigr)
\le\|x\|_2^2+\|\mathsf H_\ell x\|_2^2
\le5\|x\|_2^2.
\]
Thus \(\|A\|_{\mathrm{op}}\le\sqrt5\), even without the bounded-bulk assumption.

For the lower bound, let \(L_r\) be the \(r\times r\) lower triangular matrix of ones. For any
nonnegative diagonal matrix \(\Lambda\),
\[
2x^\top(L_r+\Lambda)x
=\|x\|_2^2+(\mathbf1_r^\top x)^2+2x^\top\Lambda x
\ge\|x\|_2^2.
\]
Cauchy--Schwarz therefore gives
\[
\sigma_{\min}(L_r+\Lambda)\ge\tfrac12,
\qquad
\|(L_r+\Lambda)^{-1}\|_{\mathrm{op}}\le2.
\]
Fix \(1\le j\le\ell\), and let \(C_j\) be the leading
\(j\times j\) principal submatrix of \(A\). Write
\[
C_j=D_j^{-1}M_j,
\qquad
D_j=\operatorname{diag}(a_i+i-1)_{i=1}^j,
\qquad
M_j=L_j+\operatorname{diag}(a_i-1)_{i=1}^j.
\]
The preceding coercivity estimate gives \(\|M_j^{-1}\|_{\mathrm{op}}\le2\).
Moreover,
\[
D_j-M_j
=\operatorname{diag}(i-1)_{i=1}^j-(L_j-I_j)=:K_j.
\]
The absolute row sums of \(K_j\) are at most \(2(j-1)\), and every
absolute column sum equals \(j-1\). Hence the Schur test gives
\(\|K_j\|_{\mathrm{op}}\le\sqrt2(j-1)\), and
\[
\|C_j^{-1}\|_{\mathrm{op}}
=\|I_j+M_j^{-1}K_j\|_{\mathrm{op}}
\le1+2\sqrt2(j-1)\le3j.
\]
For vectors supported on the first \(j\) coordinates, projecting \(Ax\)
onto its first \(j\) coordinates gives \(C_jx\). The max-min principle
therefore yields
\[
\sigma_j(A)\ge\sigma_{\min}(C_j)
\ge\frac{1}{3j},
\qquad1\le j\le\ell.
\]
Together with the operator bound, this proves the universal estimates
for every sequence length, using only nonnegative scores.

Now assume spectral separation as in Assumption~\ref{ass:spectral-separation}.
Split after the first \(k\) rows and columns, interpreting zero-size blocks
in the usual way:
\[
A=\begin{pmatrix}A_{11}&0\\A_{21}&A_{22}\end{pmatrix},
\qquad
A_{21}=z^{(b)}\mathbf1_k^\top,
\qquad
z^{(b)}=(z_{k+1},\ldots,z_\ell)^\top.
\]

Deleting the last \(\ell-k\) columns leaves a matrix of rank at most \(k\).
Lemma~\ref{lem:rank_perturbation_minmax} consequently gives
\[
\sigma_{k+m}(A)\le\sigma_m(A_{22}),
\qquad1\le m\le\ell-k.
\]

\paragraph{Step 2: spike subspace.}
For \(k\ge1\), the spike assumption gives
\(z_i\le1/(\ell f(\ell))\) for \(i\le k\).
Thus \(A_{11}=I_k+E_\ell\), where the absolute row sums and column sums
of \(E_\ell\) are at most \(2k/(\ell f(\ell))\) and
\(k/(\ell f(\ell))\), respectively. The Schur test yields
\[
\delta_\ell:=\|E_\ell\|_{\mathrm{op}}
=O\!\left(\frac{k}{\ell f(\ell)}\right)=o(1).
\]
On the \((k-1)\)-dimensional subspace
\[
E_{\mathrm{sp}}=
\{(u,0):u\in\mathbb R^k,\ \mathbf1_k^\top u=0\},
\]
the coupling block vanishes, so \(A(u,0)=(A_{11}u,0)\). Consequently,
\[
(1-\delta_\ell)\|u\|_2
\le\|A(u,0)\|_2
\le(1+\delta_\ell)\|u\|_2.
\]
This proves the almost-isometry assertion. In fact, on the whole first
\(k\)-coordinate subspace, \(\|A(u,0)\|_2\ge\|A_{11}u\|_2\), and hence
the max-min principle gives the slightly stronger conclusion
\[
\sigma_k(A)\ge1-\delta_\ell\qquad(k\ge1).
\]

\paragraph{Step 3: singular-value degeneration without bounded bulk.}
Here we use only spectral separation. Let \(P_k=\operatorname{diag}(I_k,0)\) and put
\(t_\ell=\ell/g(\ell)\ge1\) for all sufficiently large \(\ell\).
For a bulk row \(i>k\),
\[
\sum_j A_{ij}^2
=\frac{i-1+a_i^2}{(a_i+i-1)^2}
\le\frac1i+\min\!\left\{1,\frac{t_\ell^2}{i^2}\right\}.
\]
Splitting the sum at \(\lceil t_\ell\rceil\) gives
\[
\sum_{i=1}^\ell\min\!\left\{1,\frac{t_\ell^2}{i^2}\right\}
\le3t_\ell.
\]
Writing \(\mathcal H_n=\sum_{i=1}^n i^{-1}\) and \(\mathcal H_0=0\),
the block comparison above therefore yields
\[
\#\{i>k:\sigma_i(A)>\varepsilon\}
\le\frac{\mathcal H_\ell+3\ell/g(\ell)}{\varepsilon^2}.
\]
For a spike row \(i\le k\),
\[
\sum_j|(A-P_k)_{ij}|^2
=\frac{i(i-1)}{(a_i+i-1)^2}
\le\frac{i^2}{\ell^2f(\ell)^2}.
\]
Consequently,
\[
\frac{\|A-P_k\|_{\mathrm F}^2}{\ell}
\le\frac{k^3}{\ell^3f(\ell)^2}
+\frac{\mathcal H_\ell}{\ell}+\frac3{g(\ell)}
=o(1).
\]
Mirsky's singular-value inequality gives
\[
\frac1\ell\sum_{i=1}^{\ell}
\bigl(\sigma_i(A)-\mathbf1_{\{i\le k\}}\bigr)^2
\le\frac{\|A-P_k\|_{\mathrm F}^2}{\ell}=o(1).
\]
The empirical singular-value distribution approaches the moving mixture
\(c_\ell\delta_1+(1-c_\ell)\delta_0\) in mean-square transport distance.
If \(c_\ell\to c\in[0,1]\), it therefore converges weakly to
\(c\delta_1+(1-c)\delta_0\).

For Steps 4 and 5, assume \(1\le a_i\le e^B\) for \(i>k\).

\paragraph{Step 4: Frobenius control of the bulk.}
The bulk-block row associated with original row \(i>k\) has squared norm
\[
\frac{i-k-1+a_i^2}{(a_i+i-1)^2}
\le\frac{i+e^{2B}}{i^2}
\le\frac{1+e^{2B}}{i}.
\]
Consequently, we obtain
\[
\|A_{22}\|_{\mathrm F}^2\le C_B(\mathcal H_\ell-\mathcal H_k)
\le C_B\log(\ell+1),
\qquad
\sigma_{k+m}(A)
\le C_B\sqrt{\frac{\mathcal H_\ell-\mathcal H_k}{m}}.
\]
In particular this proves the logarithmic bound in the theorem. For each
\(\varepsilon>0\), the same comparison gives
\[
\#\{i>k:\sigma_i(A)>\varepsilon\}
\le\frac{\|A_{22}\|_{\mathrm F}^2}{\varepsilon^2}
\le\frac{C_B\log(\ell+1)}{\varepsilon^2}.
\]
The asserted vanishing after \(m_\ell\gg\log\ell\) follows.

\paragraph{Step 5: the lower-right block bound.}
For \(j>k\), choose the \((\ell-j+1)\)-dimensional coordinate subspace
\(F_j=\operatorname{span}\{e_j,\ldots,e_\ell\}\). By triangularity, the
restriction to this subspace is the lower-right block \(T_j\). All its
nonzero entries are at most \(e^B/j\), so coordinatewise
\[
|T_jx|\le\frac{e^B}{j}L_{\ell-j+1}|x|.
\]
Using \(\|L_r\|_{\mathrm{op}}\le\|L_r\|_{\mathrm F}\le r\) and the min-max principle gives
\[
\sigma_j(A)\le\|T_j\|_{\mathrm{op}}
\le\frac{e^B(\ell-j+1)}{j}.
\]
Combining this with Steps 1 and 4 proves all upper bounds. In particular,
if \(k<\ell\), then \(\sigma_\ell(A)=\Theta_B(\ell^{-1})\).
The proof already includes \(k=0\). If \(k=\ell\), Step 2 applies to the
whole matrix, Step 1 remains valid, and there are no bulk assertions.

\end{proof}

\begin{remark}[Quantitative bounds require additional bulk control]
\label{rem:causal-unbounded-bulk-esd}
Spectral separation suffices for weak convergence of the empirical
singular-value distribution, but additional bulk control is still needed
for the logarithmic counting and rank bounds.
For example, take \(k=0\) and \(a_i=\sqrt\ell\) for all \(i\), which
satisfies spectral separation with \(g(\ell)=\sqrt\ell\).
For \(r=\lfloor\sqrt\ell/4\rfloor\), the leading principal block
\(C_r\) satisfies \(\|C_r-I_r\|_{\mathrm{op}}\le2r/\sqrt\ell\le1/2\) by the
Schur test, so \(\sigma_r(A)\ge1/2\). Thus
\[
\#\{i:\sigma_i(A)>1/3\}\ge r=\Omega(\sqrt\ell),
\qquad
\erank_2(A)\ge\srank(A)\ge r/20=\Omega(\sqrt\ell),
\]
where the stable-rank bound uses \(\|A\|_{\mathrm F}^2\ge r/4\) and
\(\|A\|_{\mathrm{op}}^2\le5\). Hence spectral separation alone does not give the
logarithmic counting and rank bounds, even though the empirical
singular-value distribution still tends to \(\delta_0\).
\end{remark}

\subsection{Remarks on causal singular values}
\label{app:causal-singular-remarks}
\begin{remark}[Why one cannot replace the bulk upper bound by \(O(1/i)\) when \(k\) is finite]
\label{rem:causal-no-one-over-i}
Under the bounded-bulk assumptions of Theorem~\ref{thm:causal_singular_values}, if \(k\) is finite, a uniform estimate
\[
\sigma_i(A) \le \frac{\widehat C_{B,k}}{i}
\qquad \text{for all } i\in\{k+1,\dots,\ell\}
\]
cannot hold in general.

Indeed, for every bulk row \(p>k\), the row of \(A\) has \(p-1\) off-diagonal entries equal to \(z_p\) and diagonal entry \(e^{s_p}z_p\). Hence
\[
\sum_{q=1}^\ell A_{pq}^2
=
\bigl((p-1)+e^{2s_p}\bigr)z_p^2
\ge
\frac{p-1}{(e^B+p-1)^2}.
\]
Therefore there exists \(\widehat c_{B,k}>0\) such that
\[
\|A\|_{\mathrm F}^2
\ge
\sum_{p=k+1}^\ell \frac{p-1}{(e^B+p-1)^2}
\ge
\widehat c_{B,k}\sum_{p=k+1}^\ell \frac{1}{p}
\ge
\frac{\widehat c_{B,k}}{2}\log \ell
\]
for all sufficiently large \(\ell\).

On the other hand, if \(\sigma_i(A) \le \widehat C_{B,k}/i\) held for all \(i\in\{k+1,\dots,\ell\}\), then
\[
\|A\|_{\mathrm F}^2
=
\sum_{j=1}^\ell \sigma_j(A)^2
\le
k\|A\|_{\mathrm{op}}^2 + \widehat C_{B,k}^2\sum_{i=k+1}^\ell \frac{1}{i^2}
\le
\widetilde C_{B,k},
\]
which contradicts the logarithmic lower bound above. Thus the factor \(\ell-i+1\) in the upper bound of Theorem~\ref{thm:causal_singular_values} cannot, in general, be removed from the ordered bulk estimate.
\end{remark}

\begin{remark}[Spike subspace versus ordered singular values]
\label{rem:causal-spike-ordered}
Step 2 does not imply
\[
|\sigma_i(A)-1| = O\!\left(\frac{k}{\ell f(\ell)}\right)
\qquad \text{for } 1\le i\le k.
\]
For \(k\ge1\), what Step 2 proves is that the restriction of \(A\) to the \((k-1)\)-dimensional spike subspace \(E_{\mathrm{sp}}\) is almost isometric:
\[
\left(1-O\!\left(\frac{k}{\ell f(\ell)}\right)\right)\|v\|_2
\le
\|Av\|_2
\le
\left(1+O\!\left(\frac{k}{\ell f(\ell)}\right)\right)\|v\|_2,
\qquad v\in E_{\mathrm{sp}}.
\]
Equivalently, the singular values of the compressed operator \(A|_{E_{\mathrm{sp}}}\) all lie in
\[
\left[1-O\!\left(\frac{k}{\ell f(\ell)}\right),\,
1+O\!\left(\frac{k}{\ell f(\ell)}\right)\right].
\]
However, these are not the same as the ordered singular values of the full matrix \(A\). The bulk block \(A_{22}\) can already have operator norm strictly larger than \(1\); for instance, for fixed \(k\) and \(s_i=0\) for \(i>k\), it is the shifted Ces\`aro matrix, whose norm is bounded away from \(1\). Therefore the top ordered singular values of \(A\) may exceed \(1\) by an \(O(1)\) amount. Projecting onto the first \(k\) output coordinates in Step 2 gives the one-sided estimate
\[
\sigma_k(A) \ge 1 - O\!\left(\frac{k}{\ell f(\ell)}\right),\qquad k\ge1,
\]
together with the universal upper bound \(\sigma_i(A)\le \|A\|_{\mathrm{op}} \le\sqrt5\).
\end{remark}

\section{Rank proxies for attention and causal attention}
\label{app:rank-proxies}

This appendix collects the rank-proxy consequences of the singular- and
eigenvalue results in the main article.  The entropy-based proxies require
more care than pointwise spectral collapse: many individually vanishing modes
can retain non-negligible total entropy.

\begin{definition}[Rank proxies and asymptotic notation]
\label{def:rank-proxies}
For a nonzero matrix \(A\in\mathbb{R}^{\ell\times\ell}\) with singular values
\(\sigma_1(A),\ldots,\sigma_\ell(A)\), define
\[
    \srank(A)
    :=
    \frac{\|A\|_{\mathrm F}^2}{\|A\|_{\mathrm{op}}^2},
    \qquad
    \erank_1(A)
    :=
    \exp\!\left\{-\sum_{j=1}^{\ell}q_j^{(1)}\log q_j^{(1)}\right\},
    \quad
    q_j^{(1)}=\frac{\sigma_j(A)}{\sum_r \sigma_r(A)},
\]
and
\[
    \erank_2(A)
    :=
    \exp\!\left\{-\sum_{j=1}^{\ell}q_j^{(2)}
    \log q_j^{(2)}\right\},
    \qquad
    q_j^{(2)}
    =\frac{\sigma_j(A)^2}{\sum_r \sigma_r(A)^2},
\]
with \(0\log 0=0\).  For the zero matrix, set these three
singular-value proxies equal to zero.  For eigenvalues
\(\lambda_1(A),\ldots,\lambda_\ell(A)\), define
\[
    \srank_\lambda(A)
    :=
    \frac{\sum_i|\lambda_i(A)|}{\max_i|\lambda_i(A)|},
    \qquad
    \erank_\lambda(A)
    :=
    \exp\!\left\{-\sum_{i=1}^{\ell}p_i\log p_i\right\},
    \quad
    p_i=\frac{|\lambda_i(A)|}{\sum_j|\lambda_j(A)|}.
\]
The eigenvalue proxies are used when the spectrum is not identically zero; by
convention, both are zero when their normalizing spectral mass vanishes.
Thus \(\srank_\lambda\) is the stated \(\ell^1/\ell^\infty\) eigenvalue
participation proxy; it is distinct from the singular-value stable rank
\(\srank\).
Throughout, \(O(\cdot)\), \(\Omega(\cdot)\), and \(\Theta(\cdot)\) refer to
\(\ell\to\infty\); their implicit constants are independent of \(\ell\), but
may depend on fixed parameters explicitly indicated in the statement.
\end{definition}

\subsection{Bidirectional attention}

Pointwise spectral collapse does not by itself control a rank proxy: the
relevant bulk mass is quadratic for stable rank and linear for the
eigenvalue participation proxy, while entropy ranks additionally depend on
bulk entropy moments.  The following result separates these requirements.

\begin{theorem}[Rank proxies for attention]
\label{thm:attention-rank-proxies}
Under Assumptions~\ref{ass:diagonal-in-law}
and~\ref{ass:spectral-separation}, retain \(A,d_i,c_\ell,\varepsilon_\ell\)
from Section~\ref{subsec:bidirectional_attention}. Using \(\alpha_\pm\) from
Theorem~\ref{thm:singular-collapse}, define
\[
a_\pm(x)\coloneqq\alpha_\pm(x)^2,\qquad
R_\ell=k+1-c_\ell.
\]
For \(p\in\{1,2\}\), define the bulk mass and entropy moment
\[
M_{p,\ell}=\sum_{i=k+1}^{\ell}d_i^p,
\qquad
E_{p,\ell}=\sum_{i=k+1}^{\ell}d_i^p\log\frac1{d_i},
\qquad
B_{p,\ell}=M_{p,\ell}+E_{p,\ell},
\]
where a summand with \(d_i=0\) is zero.  Define also
\(\Delta_{p,\ell}=\sum_{i=1}^k(1-d_i^p)=O(k/f(\ell))\).
Then
\begin{align}
\srank(A)
&=\frac{R_\ell+(1-\ell^{-1})(M_{2,\ell}-\Delta_{2,\ell})}
{a_+(c_\ell)+O(\varepsilon_\ell)},
\label{eq:attention-stable-bulk}\\
\srank_\lambda(A)
&=R_\ell+(1-\ell^{-1})(M_{1,\ell}-\Delta_{1,\ell}).
\label{eq:attention-eigen-stable-bulk}
\end{align}
In particular, the following are necessary and sufficient conditions for
the corresponding ideal-profile equivalents:
\begin{align*}
\srank(A)&=\frac{R_\ell}{a_+(c_\ell)}(1+o(1))
&&\Longleftrightarrow\quad M_{2,\ell}=o(R_\ell),\\
\srank_\lambda(A)&=R_\ell(1+o(1))
&&\Longleftrightarrow\quad M_{1,\ell}=o(R_\ell),\\
\erank_2(A)&=(k+1)(1+o(1))
&&\Longleftrightarrow\quad B_{2,\ell}=o(R_\ell),\\
\erank_1(A)&=(k+1)(1+o(1))
&&\Longleftrightarrow\quad B_{1,\ell}=o(R_\ell),\\
\erank_\lambda(A)&=(k+1)(1+o(1))
&&\Longleftrightarrow\quad B_{1,\ell}=o(R_\ell).
\end{align*}
Each equivalence is asserted separately; no logarithmic rate condition is
needed on the spike defects \(1-d_i\).
Since \(E_{p,\ell}\ge\log g(\ell)\,M_{p,\ell}\) for all sufficiently
large \(\ell\), each entropy condition \(B_{p,\ell}=o(R_\ell)\)
is equivalently \(E_{p,\ell}=o(R_\ell)\).

For comparison with the deterministic spectral profiles, let \(N_\ell^\star\) be
the nuclear mass of the ideal singular-value profile.  Explicitly,
\[
N_\ell^\star=
\begin{cases}
1, & k=0,\\
k-1+\sqrt{a_+(c_\ell)}+\sqrt{a_-(c_\ell)}, & 1\le k<\ell,\\
\ell, & k=\ell.
\end{cases}
\]
Under \(B_{1,\ell}=o(R_\ell)\), respectively
\(B_{2,\ell}=o(R_\ell)\), the entropy equivalents can also be written as
\begin{align*}
\erank_1(A)
&=N_\ell^\star\exp\!\left\{
-\frac{\sqrt{a_+(c_\ell)}\log\sqrt{a_+(c_\ell)}
+\sqrt{a_-(c_\ell)}\log\sqrt{a_-(c_\ell)}}{N_\ell^\star}
\right\}(1+o(1)),\\
\erank_2(A)
&=R_\ell\exp\!\left\{
-\frac{a_+(c_\ell)\log a_+(c_\ell)
+a_-(c_\ell)\log a_-(c_\ell)}{R_\ell}
\right\}(1+o(1)).
\end{align*}
Under \(B_{1,\ell}=o(R_\ell)\),
\[
\erank_\lambda(A)
=R_\ell\exp\!\left\{
-\frac{(1-c_\ell)\log(1-c_\ell)}{R_\ell}
\right\}(1+o(1)).
\]
Here \(0\log0=0\); at \(k=0\) the ideal profile has just one unit
singular value and one unit eigenvalue, and at \(k=\ell\) all ideal
spectral values are one.  The displayed expressions agree with these
endpoint conventions.

Some useful consequences require only simpler conditions.  If all bulk
scores are uniformly bounded, then
\[
M_{2,\ell}=O(\ell^{-1}),\qquad
B_{2,\ell}=O(\ell^{-1}\log\ell),
\]
so the stable-rank and \(\erank_2\) equivalents hold automatically.
In particular, if also \(k=O(1)\), then
\[
\srank(A)=k+1+o(1),\qquad \erank_2(A)=k+1+o(1).
\]
For any bulk satisfying Assumption~\ref{ass:spectral-separation},
\[
B_{p,\ell}=O\!\left(\frac{(\ell-k)(1+\log g(\ell))}{g(\ell)^p}\right).
\]
Thus, if \(c_\ell\to c\in(0,1]\), all the tail conditions above hold
automatically and
\[
\frac{\srank(A)}{\ell}\longrightarrow\frac{c}{a_+(c)},
\qquad
\frac{\erank_1(A)}{\ell},\quad
\frac{\erank_2(A)}{\ell},\quad
\frac{\srank_\lambda(A)}{\ell},\quad
\frac{\erank_\lambda(A)}{\ell}
\longrightarrow c.
\]
The condition \(B_{1,\ell}=o(R_\ell)\) implies all five equivalents.
It is implied by the stronger aggregate-defect condition
\[
\tau_\ell\log\ell=o(R_\ell),\qquad
\tau_\ell=\sum_{i=1}^k(1-d_i)+\sum_{i=k+1}^{\ell}d_i,
\]
but does not impose that condition's logarithmic penalty on the spikes.
Here \(\tau_\ell=O(k/f(\ell)+(\ell-k)/g(\ell))\).
\end{theorem}

\begin{proof}
Let \(J=\ell^{-1}\mathbf1\mathbf1^\top\) and \(P=I-J\).
By~\eqref{eq:attention-decomposition},
\[
A=J+\operatorname{diag}(d_1,\ldots,d_\ell)P.
\]
Together with \(PJ=0\), this gives
\begin{equation}
\label{eq:attention-frobenius-rank-proxy}
\|A\|_{\mathrm F}^2
=1+(1-\ell^{-1})\sum_{i=1}^{\ell}d_i^2,
\end{equation}
\begin{equation}
\label{eq:attention-trace-rank-proxy}
\operatorname{tr}(A)
=1+(1-\ell^{-1})\sum_{i=1}^{\ell}d_i.
\end{equation}
Indeed, the two matrix terms are Frobenius orthogonal and the diagonal
entries of the projection \(P\) are \(1-\ell^{-1}\).
Spectral separation gives, uniformly,
\[
0\le1-d_i\le f(\ell)^{-1}\quad(i\le k),\qquad
0\le d_i\le g(\ell)^{-1}\quad(i>k).
\]
If \(d_i^\star=\mathbf1_{\{i\le k\}}\) and
\(A^\star=J+\operatorname{diag}(d^\star)P\), then
\[
\|A-A^\star\|_{\mathrm{op}}\le\varepsilon_\ell.
\]
The ideal singular profile in Theorem~\ref{thm:singular-collapse}
therefore gives
\(\|A\|_{\mathrm{op}}^2=a_+(c_\ell)+O(\varepsilon_\ell)\), including
\(k=0,\ell\).  This proves~\eqref{eq:attention-stable-bulk}.

By Theorem~\ref{thm:eigen-collapse}, all eigenvalues are nonnegative and
the largest is \(1\).  Hence
\(\srank_\lambda(A)=\operatorname{tr}(A)\), proving
\eqref{eq:attention-eigen-stable-bulk}.  Since
\(\Delta_{p,\ell}=o(R_\ell)\), the two stable-rank equivalences follow.

We next control the entropy without multiplying the spike error by
\(\log\ell\).  Write \(D=\operatorname{diag}(d)\). By
\eqref{eq:attention-decomposition}, \(A\) is a rank-one perturbation of \(D\).
The singular-value min--max principle and Hermitian interlacing give,
respectively,
\begin{equation}
\label{eq:attention-rank-tail-interlacing}
d_{j+1}\le\sigma_j(A)\le d_{j-1}\quad(2\le j\le\ell),
\qquad
d_j\le\lambda_j(A)\le d_{j-1}\quad(2\le j\le\ell),
\end{equation}
where \(d_{\ell+1}=0\).  For example, adding a rank-one matrix can
shift a singular-value index by at most one; applying that fact both to
\(A=D+z\mathbf1^\top\) and to \(D=A-z\mathbf1^\top\) proves the first
pair of inequalities.

Suppose first that \(k<\ell\), and set \(h=k+1\).  For
\(p\in\{1,2\}\), apply~\eqref{eq:attention-rank-tail-interlacing} to
the tail \(j>h\).  Both functions \(x^p\) and
\(x^p\log(1/x)\) are increasing on a sufficiently small fixed
neighborhood of zero.  The upper and lower sums obtained by interlacing
differ from the bulk sums by at most three endpoint terms, each tending
to zero.  Consequently,
\begin{align*}
\sum_{j>h}\sigma_j(A)^p&=M_{p,\ell}+o(1),&
\sum_{j>h}\sigma_j(A)^p\log\frac1{\sigma_j(A)}
&=E_{p,\ell}+o(1),\\
\sum_{j>h}\lambda_j(A)&=M_{1,\ell}+o(1),&
\sum_{j>h}\lambda_j(A)\log\frac1{\lambda_j(A)}
&=E_{1,\ell}+o(1).
\end{align*}
These conclusions include \(k=0\) and \(k=\ell-1\), with empty
tail sums interpreted as zero.  The displayed \(o(1)\) terms are
uniformly dominated by the neighboring bulk endpoint terms; if a bulk
endpoint vanishes exactly, the corresponding comparison is exact.

For the head singular values, Weyl's inequality gives \(k-1\) values
equal to \(1+O(\varepsilon_\ell)\), together with the two exceptional
ideal-profile values perturbed by \(O(\varepsilon_\ell)\); when
\(k=0\), there is only one head value, \(1+O(\varepsilon_\ell)\).
For the head eigenvalues, \(\lambda_1=1\), the next \(k-1\) values
are \(1+O(f(\ell)^{-1})\), and the intermediate value is
\(1-c_\ell+O(\varepsilon_\ell)\).
Thus the total head mass is \(R_\ell+o(R_\ell)\), for each of the
three choices of spectral weights
\[
x_j=\sigma_j(A),\qquad x_j=\sigma_j(A)^2,
\qquad x_j=\lambda_j(A).
\]
Here the ideal nuclear mass \(N_\ell^\star\) satisfies
\(N_\ell^\star/R_\ell\to1\).  Also, the head moment
\(\sum_{j\le h}x_j\log x_j\) is \(o(R_\ell)\): unit-profile
coordinates contribute \(O(k\varepsilon_\ell)\), and the exceptional
coordinates contribute \(o(R_\ell)\).  To see the last assertion
uniformly, their moments are bounded if \(k\to\infty\), while if
\(k\) is bounded then \(c_\ell\to0\) and every exceptional nonzero
head value tends to one.  If \(k=\ell\), all spectral values are
\(1+o(1)\) uniformly and the same head conclusions hold with an empty
tail.  These arguments apply along subsequences and hence also when the
endpoint regimes vary with \(\ell\).

For any of the three weight lists, let \(p=1,2,1\), respectively, and
let \(\mathcal T_\ell=\sum_jx_j\).  The preceding estimates state
\begin{equation}
\label{eq:attention-entropy-moment-reduction}
\mathcal T_\ell=R_\ell+M_{p,\ell}+o(R_\ell),
\qquad
\sum_jx_j\log x_j=-pE_{p,\ell}+o(R_\ell).
\end{equation}
For eigenvalue weights the second identity uses \(p=1\).
The normalization identity for entropy is exact:
\begin{align*}
H\!\left(\frac{x}{\mathcal T_\ell}\right)-\log R_\ell
&=\log\frac{\mathcal T_\ell}{R_\ell}
-\frac{\sum_jx_j\log x_j}{\mathcal T_\ell}\\
&=\log\frac{\mathcal T_\ell}{R_\ell}
+\frac{pE_{p,\ell}}{\mathcal T_\ell}+o(1).
\end{align*}
This identity accounts for the normalization before taking a limit;
in particular it introduces no extra factor \(\log R_\ell\).
Since \(M_{p,\ell},E_{p,\ell}\ge0\) and
\(\mathcal T_\ell/R_\ell=1+M_{p,\ell}/R_\ell+o(1)\), its right-hand
side tends to zero if and only if
\(M_{p,\ell}+E_{p,\ell}=o(R_\ell)\).
Indeed, convergence forces first \(M_{p,\ell}/R_\ell\to0\) and
then \(E_{p,\ell}/R_\ell\to0\).  Because
\(R_\ell/(k+1)\to1\), this proves all three entropy equivalences.

The ideal singular, squared-singular, and eigenvalue lists have masses
\(N_\ell^\star,R_\ell,R_\ell\), respectively.  Substitution in the entropy
definition yields the displayed deterministic-profile expressions.
Each ideal head entropy moment divided by its mass tends to zero by
the same head argument, so these expressions are equivalent to the
already proved \((k+1)(1+o(1))\) formulas.

Finally, bounded bulk scores give \(d_i=O(\ell^{-1})\) for \(i>k\),
and hence the stated quadratic bounds.  More generally,
\(d_i\le g(\ell)^{-1}\) and monotonicity near zero give the bounds
on \(B_{p,\ell}\).  Since \(B_{2,\ell}\le B_{1,\ell}\), the linear
entropy condition implies all the others.  To compare with the
aggregate-defect condition, write \(m=M_{1,\ell}\) and \(n=\ell-k\).
Concavity gives
\(E_{1,\ell}\le m\log(n/m)\) when \(m>0\) and \(n>0\).
For \(m\ge1\), this is at most \(m\log\ell\).  For \(0<m<1\),
it is at most \(m\log\ell+m\log(1/m)\); the latter term is
\(o(R_\ell)\) under \(\tau_\ell\log\ell=o(R_\ell)\), either
because \(R_\ell\to\infty\) or, along bounded subsequences, because
\(m\to0\).  This proves the final implication.
\end{proof}

\begin{remark}[Why the tail conditions differ]
Bounded bulk scores suffice for stable rank and squared-singular-value
entropy rank, but not for linear spectral mass or entropy.  For example,
take \(k=0\) and \(s_i=b>0\) for every \(i\).  Then
\[
A=dI+(1-d)J,\qquad
d=\frac{e^b-1}{e^b+\ell-1}\sim\frac{e^b-1}{\ell}.
\]
Its singular values and eigenvalues are \(1,d,\ldots,d\), so
\[
\srank(A),\ \erank_2(A)\longrightarrow1,\qquad
\srank_\lambda(A)\longrightarrow e^b,\qquad
\erank_1(A)=\erank_\lambda(A)\asymp\ell^{\,1-e^{-b}}.
\]
Even vanishing bulk mass does not suffice for entropy convergence.
For any fixed \(t>0\) and \(p\in\{1,2\}\), take \(k=0\) and
\[
d_i=d_\ell=\left(\frac{t}{\ell\log\ell}\right)^{1/p},\qquad
s_i=\log\frac{1+(\ell-1)d_\ell}{1-d_\ell}.
\]
These scores satisfy spectral separation.  The same matrix formula gives
\(M_{p,\ell}=t/\log\ell\to0\), but
\(pE_{p,\ell}\to t\), and therefore
\[
\erank_p(A)\longrightarrow e^t.
\]
For \(p=1\), \(\srank_\lambda(A)\to1\); for \(p=2\),
\(\srank(A)\to1\).  Thus the logarithmic bulk entropy moments cannot
in general be replaced by the corresponding masses alone.
\end{remark}

\subsection{Causal attention}

For causal attention, bounded bulk scores produce harmonic squared-singular
mass.  The stable ranks and the squared-singular-value entropy rank have
the same order, whereas the singular-value and eigenvalue entropy ranks
can be larger.  Throughout this subsection the causal mask is
\[
M_{ij}=\begin{cases}-\infty,&j>i,\\0,&j\le i.\end{cases}
\]

\begin{lemma}[Schatten bounds for the finite Hardy matrix]
\label{lem:hardy-schatten}
Let \(\mathsf H_n=(i^{-1}\mathbf1_{\{j\le i\}})_{i,j=1}^n\).
For every fixed \(p>1\),
\[
\sum_{j=1}^n\sigma_j(\mathsf H_n)^p\le C_p\log(n+1),
\qquad
\|\mathsf H_n\|_* = \Theta\bigl((\log(n+1))^2\bigr),
\]
where \(\|\cdot\|_*\) denotes the nuclear norm.
\end{lemma}

\begin{proof}
Partition \(\{1,\ldots,n\}\) into the nonempty dyadic intervals
\(I_r=\{2^r,\ldots,\min(2^{r+1}-1,n)\}\),
\(0\le r\le R=\lfloor\log_2 n\rfloor\).
Let \(B_d\) retain exactly the blocks \(I_r\times I_{r-d}\),
\(d\le r\le R\), of \(\mathsf H_n\), with all other entries zero.
Then \(\mathsf H_n=\sum_{d=0}^R B_d\).
For \(d\ge1\), each retained block has rank one and singular value
at most
\[
\left(\sum_{i\in I_r}i^{-2}\right)^{1/2}|I_{r-d}|^{1/2}
\le 2^{-d/2}.
\]
The retained blocks have disjoint row and column supports.  Consequently,
with \(\|X\|_{S_p}=(\sum_j\sigma_j(X)^p)^{1/p}\),
\[
\|B_d\|_{S_p}\le(R+1)^{1/p}2^{-d/2}.
\]

For \(d=0\), reindexing within a dyadic block of size \(m=|I_r|\) gives
\(D_rL_m\), where \(D_r=\operatorname{diag}(i^{-1}:i\in I_r)\) and
\(L_m=(\mathbf1_{\{q\le u\}})_{u,q=1}^m\).  The cumulative matrix has
singular values
\[
\sigma_j(L_m)
=\frac{1}{2\sin((2j-1)\pi/(4m+2))}
\asymp\frac{m}{j},\qquad 1\le j\le m.
\]
Indeed, \(L_m^{-1}\) is lower bidiagonal with diagonal entries one
and subdiagonal entries minus one.  Its Gram matrix has diagonal
\((2,\ldots,2,1)\), off-diagonal entries minus one, and eigenvalues
\(4\sin^2((2j-1)\pi/(4m+2))\), as follows by substituting the vectors
with coordinates \(\sin(u(2j-1)\pi/(2m+1))\).
Since \(\|D_r\|_{\mathrm{op}}\le2^{-r}\) and \(m\le2^r\), ideal-norm
monotonicity gives
\(\|D_rL_m\|_{S_p}^p\le C_p\sum_{j\ge1}j^{-p}<\infty\)
uniformly in \(r\).
Thus \(\|B_0\|_{S_p}\le C_p(R+1)^{1/p}\).
The Schatten triangle inequality and the convergent series
\(\sum_{d\ge1}2^{-d/2}\) prove the first assertion.

For \(p=1\), the same argument gives
\(\|B_0\|_*\le C\sum_{r=0}^R(r+1)\) and
\(\sum_{d\ge1}\|B_d\|_*\le C(R+1)\), proving the upper bound.
Pinching onto the diagonal blocks contracts the nuclear norm: the
pinched matrix is the average of the conjugations by all block-diagonal
sign matrices.  Every full dyadic block has nuclear norm at least
\(c(r+1)\), because its row multipliers are at least \(2^{-r-1}\) and
\(\|L_{2^r}\|_*\ge c2^r(r+1)\).
Summing the full blocks \(r<R\), and adjusting constants for bounded
\(n\), proves the lower bound.
\end{proof}

\begin{theorem}[Rank proxies for causal attention]
\label{thm:causal-rank-proxies}
Under Assumptions~\ref{ass:diagonal-in-law}
and~\ref{ass:spectral-separation}, let
\(A=\operatorname{softmax}(\Sigma+M)\), and suppose that
\(s_i\le B\) for all \(i>k\), where \(B\) is independent of \(\ell\).
Define
\[
L_{\ell,k}=\sum_{i=k+1}^{\ell}\frac1i,
\qquad
J_{\ell,k}=\sum_{i=k+1}^{\ell}\frac{\log i}{i},
\qquad
T_{\ell,k}=k+L_{\ell,k},
\]
with empty sums equal to zero, and set
\[
\rho_i=\frac{\exp(2s_i)+i-1}{(\exp(s_i)+i-1)^2},
\qquad W_\ell=\sum_{i=1}^{\ell}\rho_i=\|A\|_{\mathrm F}^2,
\qquad N_\ell=\|A\|_*.
\]
Then
\[
W_\ell=T_{\ell,k}
+O_B\!\left(1+\frac{k^2}{\ell f(\ell)}\right)
=(1+o(1))T_{\ell,k},
\qquad
1\le\|A\|_{\mathrm{op}}\le\sqrt5,
\]
and
\[
\srank(A),\ \erank_2(A)=\Theta_B(T_{\ell,k}),
\qquad
N_\ell=\Theta_B\bigl(k+(\log(\ell+1))^2\bigr).
\]
In particular, the singular-value entropy proxies satisfy
\[
\erank_1(A)\ge\frac{N_\ell^2}{W_\ell}
\ge c_B\frac{\bigl(k+(\log(\ell+1))^2\bigr)^2}{T_{\ell,k}},
\qquad
\erank_2(A)\le\erank_1(A)\le\ell.
\]
An additional explicit upper bound is
\[
\erank_2(A)\le W_\ell\exp\!\left\{\frac{K_\ell}{W_\ell}\right\},
\qquad
K_\ell:=\sum_{i=1}^{\ell}\rho_i\log\frac1{\rho_i}
    =J_{\ell,k}+O_B\!\left(1+\frac{k^2}{\ell f(\ell)}\right).
\]

For the eigenvalue proxies, set
\[
U_{\ell,k}=\sum_{i=k+1}^{\ell}\frac{\exp(s_i)}i,
\qquad
V_{\ell,k}=\sum_{i=k+1}^{\ell}\frac{\exp(s_i)}i
\log\!\left(\frac{i}{\exp(s_i)}\right),
\]
and define the exact eigenvalue mass and log-moment
\[
\Lambda_\ell=\sum_{i=1}^{\ell}
\frac{\exp(s_i)}{\exp(s_i)+i-1},
\qquad
Q_\ell=\sum_{i=1}^{\ell}
\frac{\exp(s_i)}{\exp(s_i)+i-1}
\log\!\left(1+\frac{i-1}{\exp(s_i)}\right).
\]
They satisfy
\[
\Lambda_\ell=k+U_{\ell,k}
+O_B\!\left(1+\frac{k^2}{\ell f(\ell)}\right),
\qquad
Q_\ell=V_{\ell,k}
+O_B\!\left(1+\frac{k^2}{\ell f(\ell)}\right),
\]
and
\[
\srank_\lambda(A)=\Lambda_\ell=\Theta_B(T_{\ell,k}),
\qquad
\erank_\lambda(A)
=\Lambda_\ell\exp\!\left\{\frac{Q_\ell}{\Lambda_\ell}\right\}.
\]
Consequently,
\[
\srank(A),\ \srank_\lambda(A),\ \erank_2(A)
=\Theta_B\!\left(k+\log\frac{\ell+1}{k+1}\right).
\]
If \(J_{\ell,k}\le C_0T_{\ell,k}\) for a fixed constant \(C_0\), then
\(\erank_\lambda(A)=\Theta_{B,C_0}(T_{\ell,k})\).
Such a constant exists whenever \(k=\Omega((\log\ell)^2)\).
The sharper singular-value equivalents are
\[
\begin{aligned}
\erank_2(A)&=k(1+o(1))
&&\text{if }\log(\ell+1)=o(k),\\
\erank_1(A)&=k(1+o(1))
&&\text{if }(\log(\ell+1))^3=o(k).
\end{aligned}
\]
At the other extreme, when \(k=0\) and \(s_i=0\) for every \(i\),
\[
\srank(A)\sim\tfrac14\log\ell,
\qquad
\srank_\lambda(A)=\mathcal H_\ell\sim\log\ell,
\qquad
\erank_\lambda(A)\sim e^{-\gamma/2}\sqrt\ell\log\ell,
\]
where \(\mathcal H_\ell=\sum_{i=1}^{\ell}i^{-1}\) and \(\gamma\) is the
Euler--Mascheroni constant.  In this case
\(\erank_2(A)=\Theta(\log\ell)\) and
\(\erank_1(A)\ge c(\log\ell)^3\).
\end{theorem}

\begin{proof}
Write \(a_i=\exp(s_i)\) and \(z_i=(a_i+i-1)^{-1}\).
The \(i\)-th causal row has \(i-1\) entries equal to \(z_i\), diagonal
entry \(a_i z_i\), and zeros thereafter, giving the stated \(\rho_i\).
For a bulk row, direct subtraction gives
\[
\rho_i-\frac1i
=\frac{(i-1)(a_i-1)^2}{i(a_i+i-1)^2},
\qquad
0\le\rho_i-\frac1i\le\frac{C_B}{i^2}.
\]
For a spike row,
\[
0\le1-\rho_i
=\frac{(i-1)(2a_i+i-2)}{(a_i+i-1)^2}
\le\frac{Ci}{\ell f(\ell)}.
\]
Summation proves the expansion of \(W_\ell\).
Its error is \(o(T_{\ell,k})\): the spike error divided by
\(T_{\ell,k}\ge k\) is at most \(Ck/(\ell f(\ell))=o(1)\), and
\(T_{\ell,k}\to\infty\), making the bounded remainder negligible.

The exact decomposition
\[
A=D\mathsf H_\ell+C,
\qquad
D_{ii}=\frac{i}{a_i+i-1},
\qquad C_{ii}=\frac{a_i-1}{a_i+i-1},
\qquad D+C=I
\]
has diagonal matrices \(0\le D,C\le I\).
Step 1 of Appendix~\ref{app:proof-thm:causal_singular_values} gives
\(\|A\|_{\mathrm{op}}\le\sqrt5\).
Since \(A\mathbf1=\mathbf1\), its operator norm is at least one.
This proves the stable-rank bounds.

For \(h(x)=x\log(1/x)\), with \(h(0)=0\), the row estimates and the
mean-value theorem give
\[
h(\rho_i)=\frac{\log i}{i}
+O_B\!\left(\frac{\log(i+1)}{i^2}\right)\quad(i>k).
\]
For spikes, \(0\le h(1-\varepsilon)\le\varepsilon\) for
\(0\le\varepsilon\le1\), so
\[
0\le\sum_{i\le k}h(\rho_i)
\le\sum_{i\le k}(1-\rho_i)
=O\!\left(\frac{k^2}{\ell f(\ell)}\right).
\]
Their sum is the claimed expansion of \(K_\ell\).
The normalized squared singular values are the eigenvalues of
\(AA^\top/W_\ell\), whose diagonal is \((\rho_i/W_\ell)_i\).
By Schur--Horn and Schur concavity of entropy,
\[
\log\erank_2(A)\le\log W_\ell+K_\ell/W_\ell.
\]

The sharper entropy order follows from the Schatten bound.
Fix \(p\in(1,2)\).  Bounded bulk scores imply
\[
\|C\|_{S_p}^p
\le k+(e^B-1)^p\sum_{i=k+1}^{\ell}i^{-p}
\le k+C_{B,p}.
\]
The Schatten triangle inequality and Lemma~\ref{lem:hardy-schatten}
therefore give
\[
\sum_j\sigma_j(A)^p
\le C_{B,p}\bigl(k+\log(\ell+1)\bigr)
\le C_{B,p}T_{\ell,k},
\]
where \(\log(\ell+1)\le\log(k+1)+L_{\ell,k}\le T_{\ell,k}\).
For \(0<x\le1\),
\(x^2\log(1/x^2)\le C_px^p\); for \(x>1\) the left-hand side is
nonpositive.  Thus
\[
\log\erank_2(A)
=\log W_\ell+\frac1{W_\ell}
\sum_j\sigma_j(A)^2\log\frac1{\sigma_j(A)^2}
\le\log W_\ell+C_B.
\]
Conversely, Shannon entropy is at least minus the logarithm of the largest
probability, so \(\erank_2(A)\ge\srank(A)\).
This proves \(\erank_2(A)=\Theta_B(T_{\ell,k})\).

We next estimate \(N_\ell\).  Let \(P\) project onto the bulk coordinates.
On these coordinates \(D_{ii}\ge e^{-B}\); hence
\[
\|PD\mathsf H_\ell\|_*
\ge e^{-B}\|P\mathsf H_\ell\|_*
\ge e^{-B}\bigl(\|\mathsf H_\ell\|_*-2k\bigr).
\]
The last inequality uses that \((I-P)\mathsf H_\ell\) has rank at most
\(k\) and operator norm at most two.  Also
\(\|PC\|_*\le(e^B-1)\mathcal H_\ell\).
The reverse triangle inequality therefore gives
\[
N_\ell\ge\|PA\|_*
\ge c_B(\log(\ell+1))^2-C_Bk-C_B\log(\ell+1).
\]
Meanwhile, \(N_\ell\ge|\operatorname{tr}A|\ge k/2\) for all
sufficiently large \(\ell\), since spike diagonal entries tend uniformly
to one.  Use the first bound when \(k\) is a sufficiently small multiple
of \((\log(\ell+1))^2\), and the second otherwise, to obtain
\(N_\ell\ge c_B(k+(\log(\ell+1))^2)\).
The reverse bound follows from
\[
N_\ell\le\|\mathsf H_\ell\|_*+\|C\|_*
\le C_B\bigl(k+(\log(\ell+1))^2\bigr).
\]
For \(q_j^{(1)}=\sigma_j(A)/N_\ell\), Shannon entropy dominates the
order-two Renyi entropy, yielding
\(\erank_1(A)\ge1/\sum_j(q_j^{(1)})^2=N_\ell^2/W_\ell\).
Finally, for
\(q_j^{(\alpha)}=\sigma_j(A)^\alpha/\sum_r\sigma_r(A)^\alpha\),
\[
\frac{\mathrm d}{\mathrm d\alpha}H(q^{(\alpha)})
=-\alpha\operatorname{Var}_{q^{(\alpha)}}(\log\sigma_j(A))\le0.
\]
Taking \(\alpha=1,2\) proves \(\erank_2(A)\le\erank_1(A)\),
and \(\erank_1(A)\le\ell\) follows from the maximum entropy bound.

For the eigenvalue proxies, triangularity gives
\[
\lambda_i(A)=\frac{a_i}{a_i+i-1}>0,
\qquad\max_i\lambda_i(A)=\lambda_1(A)=1.
\]
The exact expressions for \(\srank_\lambda\) and
\(\erank_\lambda\) follow immediately.  For \(i>k\),
\[
\lambda_i(A)=\frac{a_i}{i}+O_B(i^{-2}),
\qquad
h(\lambda_i(A))=\frac{a_i}{i}\log\frac{i}{a_i}
+O_B\!\left(\frac{\log(i+1)}{i^2}\right).
\]
For \(i\le k\),
\[
0\le h(\lambda_i(A))\le1-\lambda_i(A)
\le\frac{i}{\ell f(\ell)}.
\]
Summation proves the expansions of \(\Lambda_\ell\) and \(Q_\ell\).
Since \(1\le a_i\le e^B\) in the bulk,
\[
U_{\ell,k}=\Theta_B(L_{\ell,k}),
\qquad
|V_{\ell,k}|\le e^B(J_{\ell,k}+BL_{\ell,k}).
\]
This gives \(\Lambda_\ell=\Theta_B(T_{\ell,k})\) and, when
\(J_{\ell,k}\le C_0T_{\ell,k}\), gives
\(Q_\ell/\Lambda_\ell=O_{B,C_0}(1)\).
The harmonic-sum estimate
\[
L_{\ell,k}\asymp\log\frac{\ell+1}{k+1}
\]
includes the empty-sum endpoint.  Also
\(J_{\ell,k}\le\log\ell\,L_{\ell,k}=O((\log\ell)^2)\),
so \(k=\Omega((\log\ell)^2)\) implies the stated condition on \(J\).

It remains to prove the two singular-value equivalents.
The spike principal block is \(I_k+E_\ell\), where the Schur test gives
\[
\delta_\ell:=\|E_\ell\|_{\mathrm{op}}
=O\!\left(\frac{k}{\ell f(\ell)}\right)=o(1).
\]
The max-min principle on the first \(k\) coordinates gives
\(\sigma_j(A)\ge1-\delta_\ell\) for \(j\le k\).
The best rank-\(k\) approximation property of Schatten norms,
applied to \((I-P)A\), gives
\[
\sum_{j>k}\sigma_j(A)^p\le\|PA\|_{S_p}^p
\le C_{B,p}\log(\ell+1),\qquad1<p<2,
\]
and
\[
\tau_\ell:=\sum_{j>k}\sigma_j(A)\le\|PA\|_*
\le C_B(\log(\ell+1))^2.
\]
Here both estimates follow from \(PA=PD\mathsf H_\ell+PC\) and
the Hardy Schatten bounds, since \(\|PC\|_{S_p}^p\le C_{B,p}\).

If \(\log(\ell+1)=o(k)\), then \(W_\ell=k+o(k)\).  Put
\[
S_1=\sum_{j\le k}\sigma_j(A)^2,
\qquad R=W_\ell-S_1,
\qquad
\Delta_-=\sum_{j\le k}\bigl(1-\sigma_j(A)^2\bigr)_+ .
\]
The spike lower bound gives
\(S_1\ge k(1-2\delta_\ell)\) and \(\Delta_-\le2k\delta_\ell\), while
\(R\le W_\ell-k+2k\delta_\ell=o(k)\).  If
\(\Delta_+=\sum_{j\le k}(\sigma_j(A)^2-1)_+\), then the identity
\[
\Delta_+=(S_1-k)+\Delta_-=(W_\ell-k)-R+\Delta_-
\]
shows that \(\Delta_+=o(k)\).  Thus the total squared mass above one,
\(\Delta_++\sum_{j>k}(\sigma_j(A)^2-1)_+\), is \(o(k)\).  The positive
part of the entropy moment (the terms with \(\sigma_j\le1\)) is bounded
by \(Ck\delta_\ell+C_B\log(\ell+1)=o(k)\), using the fixed-\(p\)
Schatten estimate above.  Since all singular values are at most
\(\sqrt5\) and \(x\log x\le C(x-1)\) on \([1,5]\), the negative
part is also \(o(k)\).  Substituting into the entropy formula proves
\(\erank_2(A)=k(1+o(1))\).

For \(\erank_1\), the row decomposition and the spike block bound give
\[
k(1-\delta_\ell)\le N_\ell
\le k(1+\delta_\ell)+C_B(\log(\ell+1))^2.
\]
The first \(k\) terms obey \(h(\sigma_j(A))\le C\delta_\ell\).
For the tail, concavity of \(h\) yields
\[
\sum_{j>k}h(\sigma_j(A))
\le\tau_\ell\log\frac{\ell-k}{\tau_\ell}
\le\tau_\ell\log\ell+\frac1e
\le C_B(\log(\ell+1))^3,
\]
with a zero tail when \(k=\ell\).
If \((\log(\ell+1))^3=o(k)\), then
\[
\log\erank_1(A)
\le\log N_\ell+
\frac{Ck\delta_\ell+C_B(\log(\ell+1))^3}{N_\ell}
=\log k+o(1).
\]
The lower bound \(\erank_1(A)\ge N_\ell^2/W_\ell=k(1+o(1))\)
proves the equivalent.

Finally, in the zero-score case \(A=\mathsf H_\ell\) and
\(W_\ell=\mathcal H_\ell\).  For the test vector \(x_j=j^{-1/2}\),
\[
\|x\|_2^2=\mathcal H_\ell,
\qquad
(Ax)_i=\frac1i\sum_{j=1}^i\frac1{\sqrt j}
=\frac2{\sqrt i}+O(i^{-1}),
\]
so \(\|Ax\|_2^2=4\mathcal H_\ell+O(1)\).
Together with \(\|\mathsf H_\ell\|_{\mathrm{op}}\le2\), this proves
\(\|A\|_{\mathrm{op}}\to2\) and the stated stable-rank equivalent.
The eigenvalues are exactly \(1/i\), whence
\[
\Lambda_\ell=\mathcal H_\ell=\log\ell+\gamma+o(1),
\qquad
Q_\ell=\sum_{i=1}^{\ell}\frac{\log i}{i}
=\tfrac12(\log\ell)^2+O(1).
\]
Thus \(Q_\ell/\Lambda_\ell=\tfrac12\log\ell-\gamma/2+o(1)\),
proving the final eigenvalue-entropy equivalent.
\end{proof}

\begin{remark}[Weaker assumptions for the order bounds]
\label{rem:causal-rank-weaker-assumptions}
The estimates
\[
\srank(A),\ \srank_\lambda(A),\ \erank_2(A)
=\Theta_{B,\eta}\!\left(k+\log\frac{\ell+1}{k+1}\right)
\]
need only \(a_i=e^{s_i}\ge1\), \(a_i\le e^B\) for \(i>k\), and
\(a_i\ge\eta i\) for \(i\le k\), where \(\eta>0\) is fixed.
Indeed, the spike diagonal entries are then at least \(\eta/(1+\eta)\),
so \(W_\ell\asymp_{B,\eta}T_{\ell,k}\); the operator and Schatten
bounds above remain valid.  The triangular eigenvalue formulas likewise
give \(\Lambda_\ell\asymp_{B,\eta}T_{\ell,k}\).
The nuclear-norm order and the resulting \(\erank_1\) lower bound also
hold, with constants depending on \(B,\eta\).
These order statements require neither ordering of the scores nor
asymptotically diverging spike separation.  The stronger equivalents in
the theorem use the vanishing spike deficits supplied by
Assumption~\ref{ass:spectral-separation}.
\end{remark}

\begin{remark}[Polynomial entropy rank of the Hardy matrix]
\label{rem:hardy-erank-one}
For zero scores, the two singular-value entropy ranks have different
growth scales:
\[
\erank_2(\mathsf H_\ell)=\Theta(\log\ell),
\qquad
\log\erank_1(\mathsf H_\ell)=\Theta(\log\ell).
\]
Thus \(\erank_1(\mathsf H_\ell)=\ell^{\Theta(1)}\), strengthening the
polylogarithmic lower bound in Theorem~\ref{thm:causal-rank-proxies}.

To see this, use the dyadic blocks in
Lemma~\ref{lem:hardy-schatten} and let \(B_0\) retain only the diagonal
blocks, with \(R=\lfloor\log_2\ell\rfloor\).
In each row block, the entries strictly before its diagonal block form
a rank-one matrix; hence
\(\operatorname{rank}(\mathsf H_\ell-B_0)\le R\).
Both matrices have operator norm at most two.  Singular-value rank
interlacing therefore implies that their counting functions
\(n_X(t)=\#\{j:\sigma_j(X)>t\}\) differ by at most \(R\).
For \(h(t)=t\log(1/t)\), with \(h(0)=0\), it follows that
\[
\left|\sum_j h(\sigma_j(\mathsf H_\ell))
-\sum_j h(\sigma_j(B_0))\right|
\le R\int_0^2|h'(t)|\,\mathrm dt=O(\log\ell).
\]
Here we used
\(\sum_jh(\sigma_j(X))=\int_0^2h'(t)n_X(t)\,\mathrm dt\).
Every full dyadic block of size \(m=2^r\) has
\(c/j\le\sigma_j\le C/j\), by the formula for
\(\sigma_j(L_m)\) and the bounds on its row multipliers.
For all sufficiently large \(j\), its entropy summand is consequently
comparable to \((\log j)/j\); the remaining finitely many summands
are bounded.  Summing first over \(j\) and then over the full blocks
\(r<R\) gives
\[
\sum_j h(\sigma_j(B_0))=\Theta((\log\ell)^3).
\]
The possibly truncated final block obeys the same upper singular-value
bound, contributes at most \(O((\log\ell)^2)\), and has a negative
part bounded by a constant.  Since
\(N_\ell=\|\mathsf H_\ell\|_*=\Theta((\log\ell)^2)\), the entropy identity
\[
\log\erank_1(\mathsf H_\ell)
=\log N_\ell+
\frac{\sum_jh(\sigma_j(\mathsf H_\ell))}{N_\ell}
\]
proves the claim.  Determining an exact exponent or a sharp equivalent
requires finer spectral information.
\end{remark}

\section{Auxiliary matrix norm bound}
\label{app:auxiliary-norm-bounds}

\begin{lemma}[Schur test]
\label{lem:schur_test}
Let \(A=(a_{pq})_{1\le p,q\le m}\) be a real \(m\times m\) matrix. Assume that
\[
\sup_{1\le p\le m}\sum_{q=1}^m |a_{pq}| \le R,
\qquad
\sup_{1\le q\le m}\sum_{p=1}^m |a_{pq}| \le C.
\]
Then
\[
\|A\|_{\mathrm{op}} \le \sqrt{RC}.
\]
In particular, if every row sum and every column sum of \(|A|\) are bounded by \(M\), then
\[
\|A\|_{\mathrm{op}} \le M.
\]
\end{lemma}


\begin{lemma}[Hardy inequality on \(\ell^2\)]
\label{lem:hardy_l2}
For \(m\ge 1\) and \(u\in\mathbb{R}^m\), define
\[
(\mathsf H_m u)_p=\frac{1}{p}\sum_{q=1}^p u_q,
\qquad 1\le p\le m.
\]
Then
\[
\sum_{p=1}^m
\left(\frac{1}{p}\sum_{q=1}^p |u_q|\right)^2
\le
4\sum_{q=1}^m |u_q|^2,
\qquad\text{equivalently}\qquad
\|\mathsf H_m|u|\|_2\le 2\|u\|_2.
\]
Moreover, for any integer \(k\ge 1\), the shifted Hardy operator
\[
(\mathsf H_{m,k}^{\mathrm{shift}} u)_p=\frac{1}{k+p-1}\sum_{q=1}^p u_q,
\qquad 1\le p\le m,
\]
satisfies
\[
\|\mathsf H_{m,k}^{\mathrm{shift}}|u|\|_2\le 2\|u\|_2.
\]
Consequently, if a linear map \(M:\mathbb{R}^m\to\mathbb{R}^m\) satisfies
\[
|(Mv)_p|\le C(\mathsf H_{m,k}^{\mathrm{shift}}|v|)_p
\qquad \text{for all }1\le p\le m,
\]
then \(\|M\|_{\mathrm{op}}\le 2C\).
\end{lemma}


\begin{lemma}[Rank perturbation and singular values]
\label{lem:rank_perturbation_minmax}
Let \(M=N+R\) be real \(\ell\times \ell\) matrices, and suppose
\(\operatorname{rank}(N)\le k\). Then, for every \(1\le m\le \ell-k\),
\[
\sigma_{k+m}(M)\le \sigma_m(R).
\]
\end{lemma}


\section{Additional experimentation details}
\label{app:additional-experimentation-details}
\label{app:empirical-scope}

The hidden-state angular diagnostics and GPT-2 score-rank summaries
(Figures~\ref{fig:wikitext-distance-metrics}, \ref{fig:vit-layer-distance-boxplot},
and~\ref{fig:score-rank-wikitext}) were performed on an AMD EPYC 9554
dual-socket CPU server with 64 cores per socket and 126 GB of RAM, without
GPU acceleration, requiring approximately 24 hours of wall-clock time in total.
The real-versus-canonical comparisons and score-level angular and spectral
diagnostics (Sections~\ref{subsec:real-canonical-comparison}
and~\ref{subsec:score-geometry}) were performed on a MacBook Pro with an
M1 Pro chip and 16 GB of unified memory, requiring approximately two hours
in total. The reported timings are totals for each experiment group.

The analysis code and saved figure inputs are released at
\url{https://github.com/Mengda-Li/attention}, with the authors' code under
the MIT license. The repository README and
\texttt{docs/REPRODUCING\_PAPER.md} provide environment setup,
data preparation, and a figure-to-script guide; \texttt{paper\_figures/README.md}
documents the publication figures and their inputs and output definitions.
The guides distinguish rendering from supplied CSV or archived figure inputs
from collecting model measurements and record the available collection
workflows and their limitations.

\subsection{Licenses for existing assets}
\label{app:asset-licenses}

All existing assets are used only for non-commercial research evaluation and
spectral diagnostics, and we do not redistribute the underlying datasets or
model weights.  The datasets used in the experiments are Imagenette
\citep{Imagenette}, a subset of ImageNet \citep{imagenet}, and Wikitext-103
\citep{merity2017pointer}.  ImageNet states that it does not own the copyright
of the images and provides access under terms permitting non-commercial research
and educational use; these terms are available at
\url{https://image-net.org/accessagreement}.  The Wikitext-103 data were
accessed through the Hugging Face dataset repository
\url{https://huggingface.co/datasets/Salesforce/wikitext}, whose repository
metadata lists the licenses as \texttt{cc-by-sa-3.0} and \texttt{gfdl}.

For language-model experiments, we use GPT-2 \citep{GPT2} through the Hugging
Face Transformers library, which is licensed under Apache-2.0
\url{https://github.com/huggingface/transformers}.  The GPT-2 weights are
loaded from \texttt{openai-community/gpt2} on Hugging Face
\url{https://huggingface.co/openai-community/gpt2}, whose model card lists the
license as MIT.  For vision experiments, we use the ViT-B/16 implementation
from TorchVision \url{https://github.com/pytorch/vision}, which is licensed
under BSD-3-Clause.  The \texttt{IMAGENET1K\_SWAG\_E2E\_V1} weights are
documented by TorchVision as SWAG-derived weights, and the SWAG repository
releases its model weights under CC-BY-NC 4.0
\url{https://github.com/facebookresearch/SWAG/blob/main/LICENSE}.

\subsection{Additional experimental figures}

\begin{figure}[htbp]
 \centering
 \includegraphics[width=\linewidth]{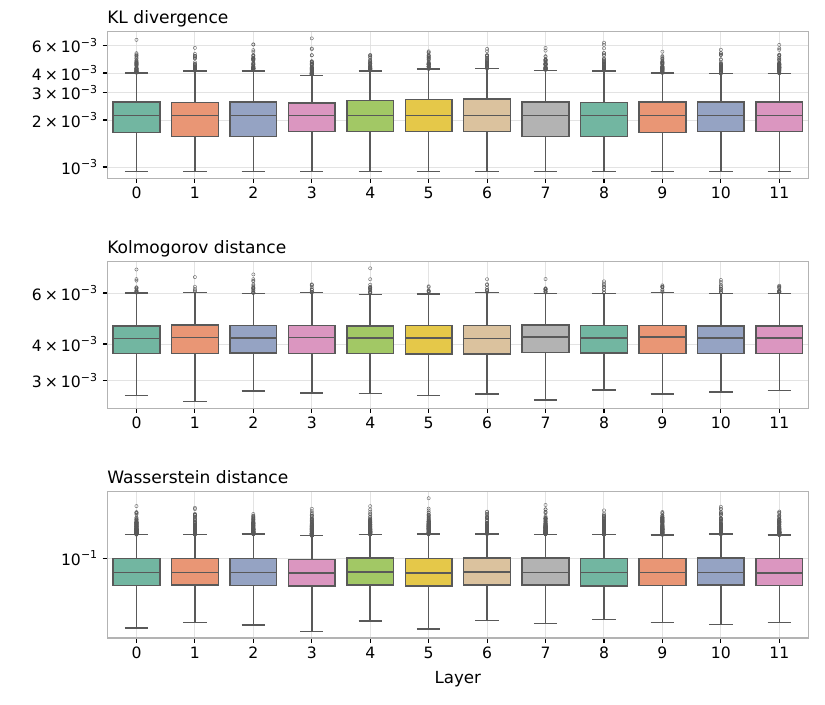}
 \caption{Layer-wise discrepancies between the empirical eigen-angle
 distribution of the ViT token-space singular matrices and the uniform
 distribution on $[-\pi,\pi]$ for Imagenette validation inputs. The log-scale
 box plots show that Kullback-Leibler divergence, Kolmogorov distance, and
 Wasserstein distance remain close to zero across layers.}
 \label{fig:vit-layer-distance-boxplot}
\end{figure}

Figures~\ref{fig:vit-layer-distance-boxplot} and~\ref{fig:score-rank-wikitext}
provide supplementary hidden-state diagnostics and score-rank summaries.

\subsection{Additional experiment details}
\label{app:real-canonical-comparison}

The direct attention comparison uses 12 WikiText-103 validation documents at
nested lengths $64,128,256,384,512,768,1024$, with all 12 GPT-2 layers and
12 heads, giving 12,096 matched comparisons. The ViT comparison uses 20
class-balanced Imagenette images, all 12 layers and 12 heads, and sequence
length 577, giving 2,880 comparisons. Sampling uses seed 233. The images
are evaluated at resolution $384\times384$ with
\texttt{IMAGENET1K\_SWAG\_E2E\_V1} weights. All matrices are analyzed per
sample, layer, and head before pooling the resulting statistics.

Attention singular values are computed by dense full SVD in float64. Score
singular values are obtained from compact QR factorizations of $Q_h$ and
$K_h$ followed by an SVD of
\(R_QR_K^\top/\sqrt{d_{qk}}\), with structural zeros appended. The core has
dimension at most 64; this reduction preserves all score singular values.
The $W_1$ comparison assigns mass $1/\ell$ to each singular value; the
singular values themselves are unnormalized. Quantile bands summarize
variation across observations, including dependent heads and nested text
prefixes.

These comparisons concern empirical spectral laws. Vanishing $W_1$ would
imply a common weak limit whenever either limit exists, while the measured
finite-length decrease provides evidence consistent with this behavior.
The equally weighted distance can be small even when a few leading modes
differ, and does not establish operator-norm or rank-proxy agreement.
For example, at GPT-2 length 1024, the pooled stable-rank medians are
1.1823 for real attention and 27.5050 for canonical attention; the
corresponding singular-value entropy-rank medians are 314.0596 and 95.1198.
The fixed-length ViT comparison describes layerwise agreement.

\subsubsection{Angular diagnostics}

For the original hidden-state diagnostics, KL divergence is computed from
50 equal-width angle bins, with $10^{-10}$ added to each bin probability
before renormalization. Kolmogorov distance compares the empirical angle CDF with
the continuous uniform CDF. The reported angular Wasserstein statistic is
the ordinary one-dimensional $W_1$ between the observed angles and an
independent uniform sample of the same size on $[-\pi,\pi]$:
\[
\widehat W_1=\frac1\ell\sum_{j=1}^{\ell}
|\theta_{(j)}-u_{(j)}|,
\qquad u_j\overset{\mathrm{iid}}\sim\operatorname{Unif}[-\pi,\pi],
\]
where parentheses denote ascending order. This uses the fixed angle
interval, without optimizing over a circular shift. It includes reference
sampling noise; finite-size effects in these diagnostics mean that a
decreasing discrepancy alone does not quantify increasing Haar isotropy.

\begin{figure}[p]
 \centering
 \includegraphics[width=\linewidth]{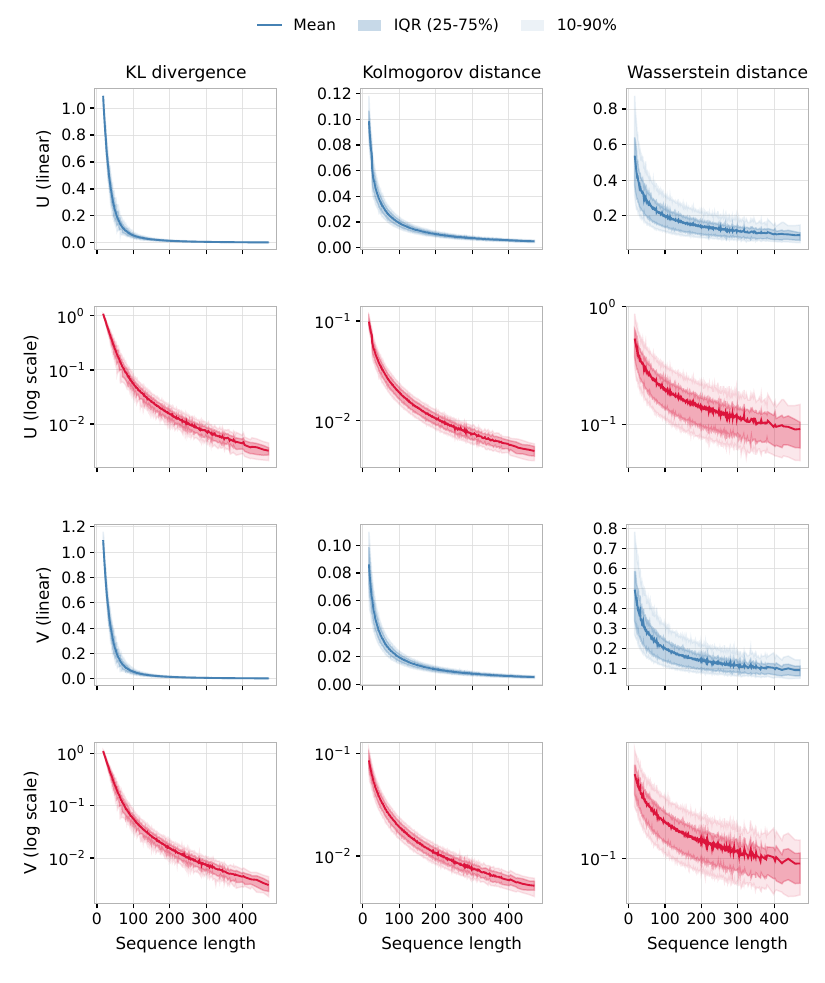}
 \caption{GPT-2 actual-score eigen-angle discrepancies for the left and
 right singular-vector matrices. KL, Kolmogorov, and angular Wasserstein
 discrepancies decrease over the displayed sequence lengths.}
 \label{fig:score-gpt2-angles}
\end{figure}

\begin{figure}[p]
 \centering
 \includegraphics[width=\linewidth]{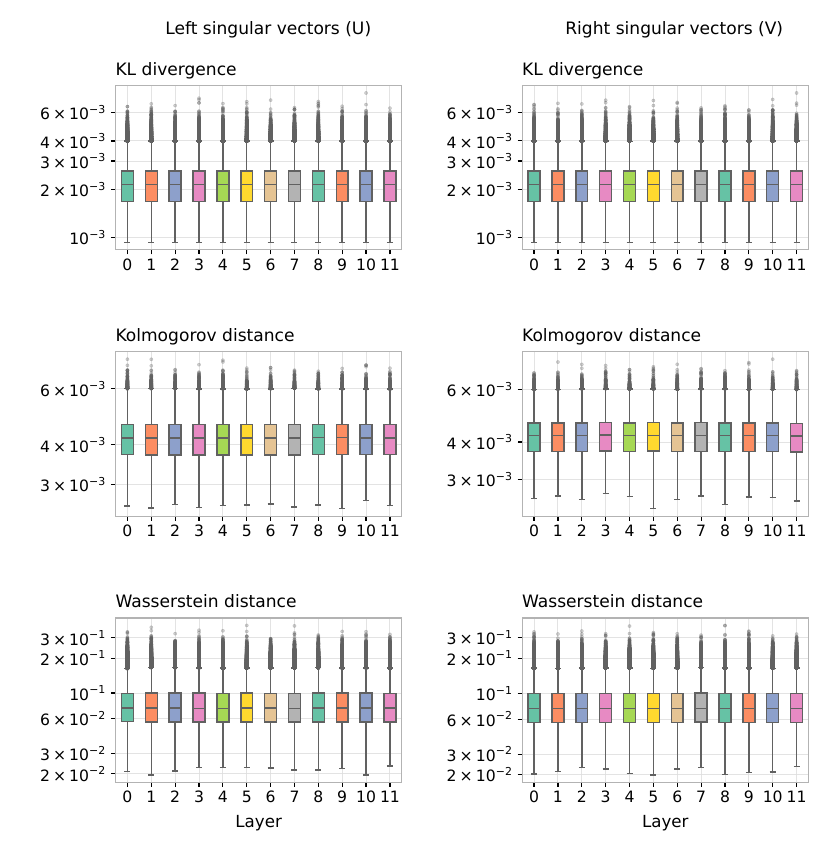}
 \caption{ViT actual-score eigen-angle discrepancies across layers for both
 singular-vector factors. Each layer contains 47,100 image--head
 observations from 3,925 images.}
 \label{fig:score-vit-angles}
\end{figure}

The score-orientation experiment covers 1,646 GPT-2 sequences of lengths
17--471 and 3,925 ViT images, producing 237,024 and 565,200 individual
score decompositions, respectively. The angular means reported in
Section~\ref{subsec:score-geometry} pool scalar discrepancies over inputs,
layers, and heads; the GPT-2 length bands 17--64 and 385--471 contain
254 and 8 sequences, respectively. Figures~\ref{fig:score-gpt2-angles}
and~\ref{fig:score-vit-angles} report eigen-angle discrepancies for the
square orthogonal SVD factors. Their interpretation depends on the
numerical SVD convention, including signs and nullspace completion.
Uniform eigen-angles are a necessary asymptotic consequence of Haar
behavior, rather than a characterization of the Haar law.

\subsection{Representative score spectra}

\begin{figure}[p]
 \centering
 \includegraphics[width=\linewidth]{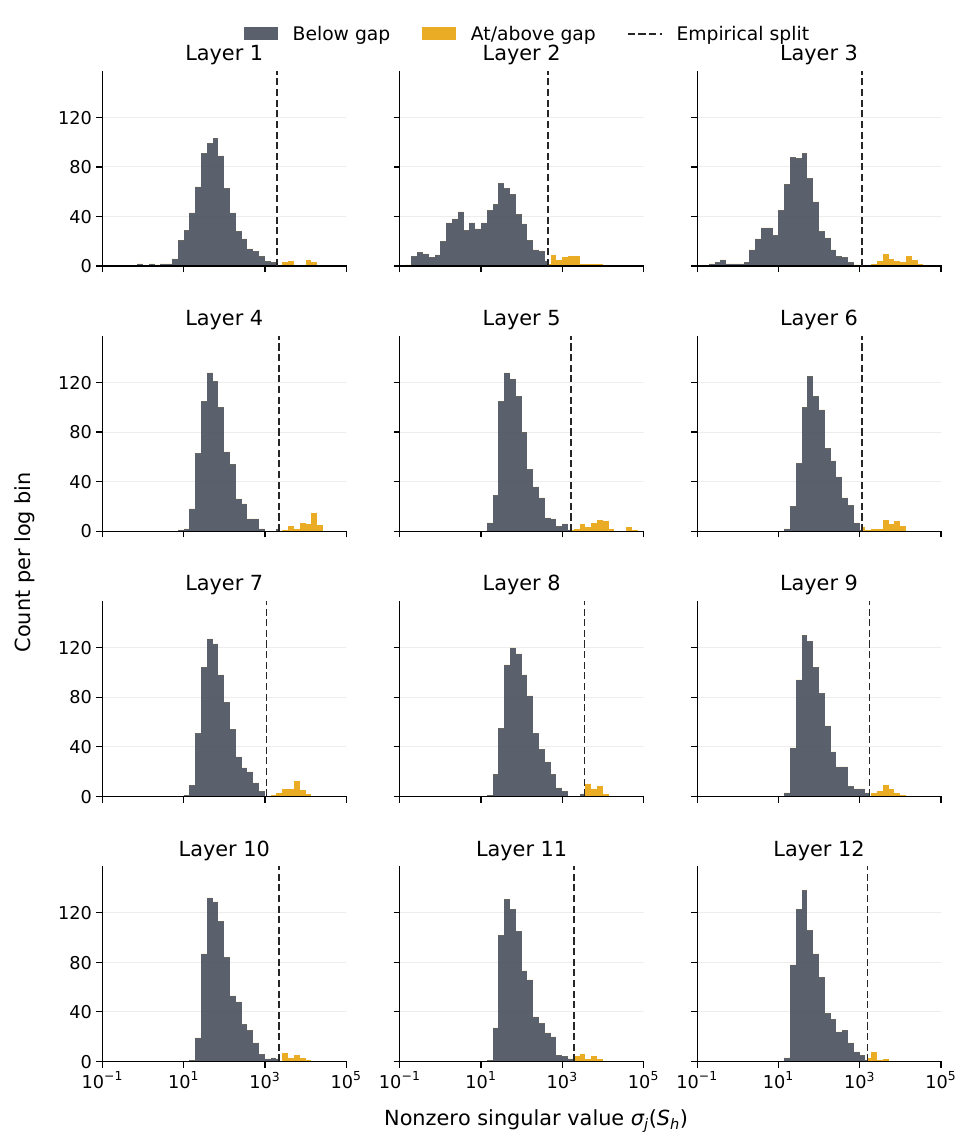}
 \caption{GPT-2 actual score spectra for a representative 1,024-token input.
 Each layer pools the nonzero singular values from all 12 heads, shown on
 a logarithmic horizontal axis. The dashed line marks the empirical gap,
 with larger values colored orange.}
 \label{fig:gpt2-score-spectrum}
\end{figure}

\begin{figure}[p]
 \centering
 \includegraphics[width=\linewidth]{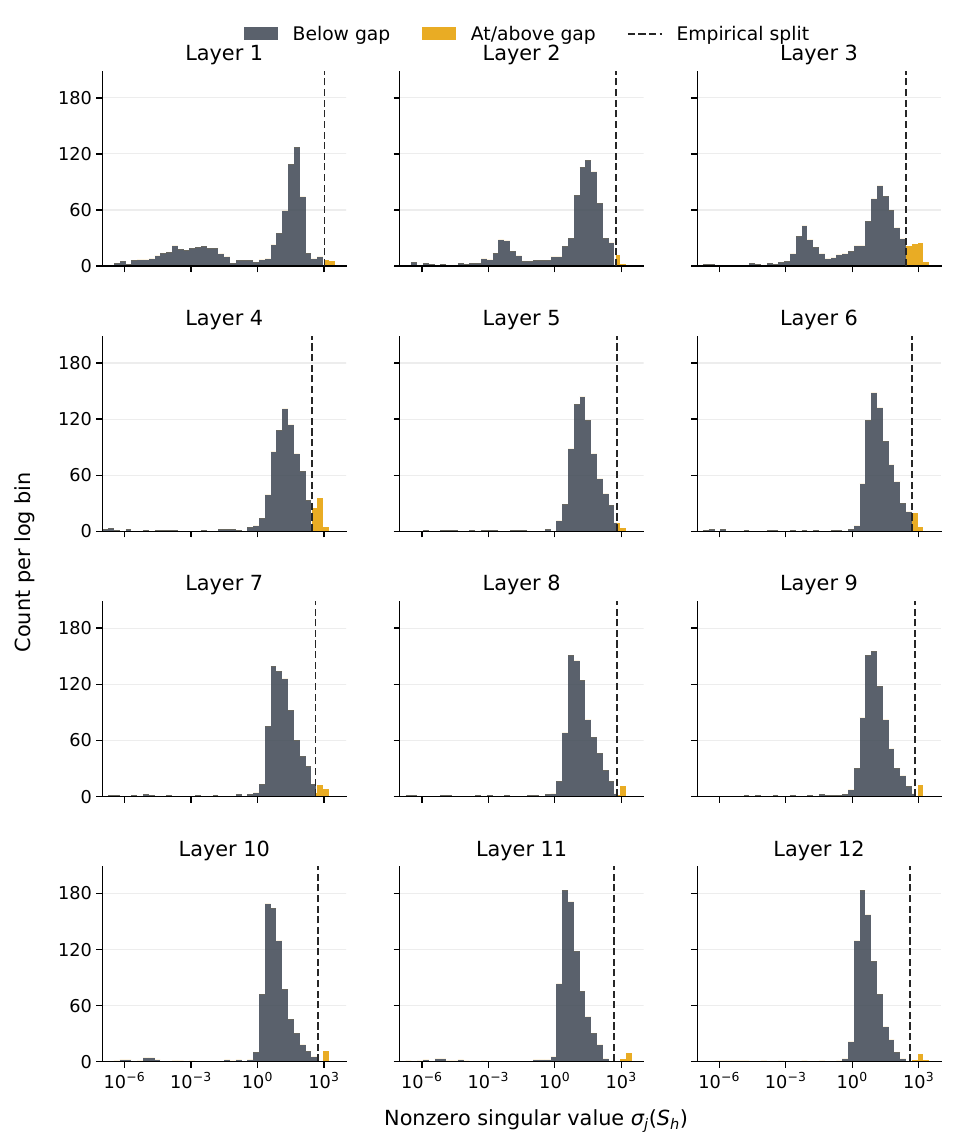}
 \caption{ViT actual score spectra for a representative 577-token image.
 Each layer pools the nonzero singular values from all 12 heads. The
 logarithmic scale and empirical-gap coloring follow
 Figure~\ref{fig:gpt2-score-spectrum}.}
 \label{fig:vit-score-spectrum}
\end{figure}

Figures~\ref{fig:gpt2-score-spectrum} and~\ref{fig:vit-score-spectrum} use
one 1,024-token GPT-2 input and one 577-token ViT input. Each layer pools
nonzero singular values from 12 independently decomposed heads; structural
zeros are omitted. The color boundary is placed at the largest adjacent
ratio in the upper half of the layer-median per-head spectrum. This
empirical split illustrates the spectrum without testing the asymptotic
thresholds in Assumption~\ref{ass:spectral-separation}. Small rank proxies
can also arise from smooth spectral decay. The representative plots and
rank summaries therefore motivate bulk--spike modeling but do not measure
how often the quantitative separation or bounded-bulk conditions hold.
The two pretrained models also differ in architecture and training, so
their comparison does not isolate the effect of causal masking.

\FloatBarrier

\subsection{Multi-head attention experiments}
\label{subsec:head-heterogeneity}
\label{app:head-heterogeneity}

\begin{figure}[htbp]
 \centering
 \includegraphics[width=\linewidth]{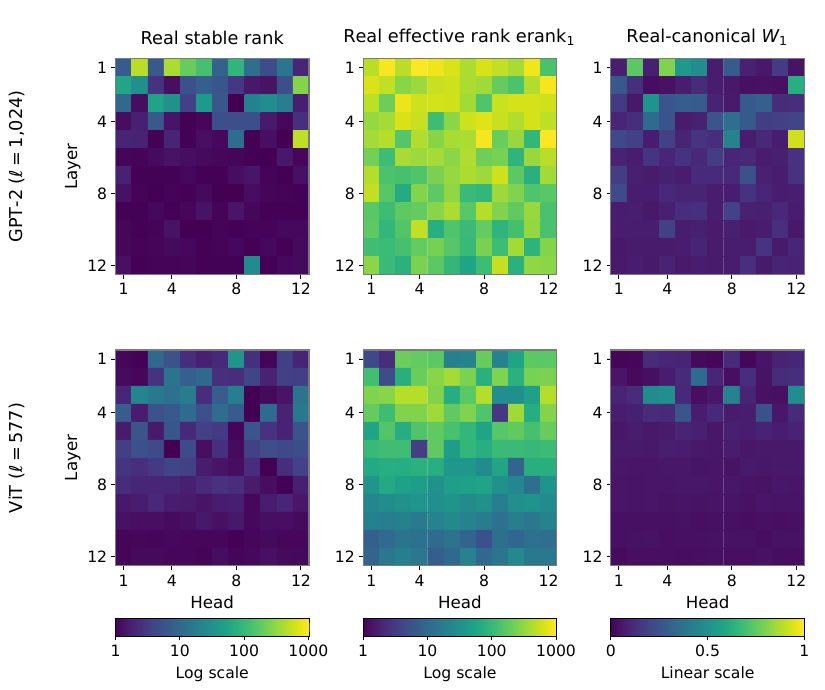}
 \caption{Head-resolved spectral diagnostics for GPT-2 (top) and ViT
 (bottom). Each cell is the median over inputs for one layer and head:
 real-attention stable rank (left), real-attention $\erank_1$ (middle),
 and real-versus-canonical spectral $W_1$ (right). Rank colors use
 logarithmic scales and $W_1$ colors a linear scale, shared across
 models within each column. Head indices follow each layer's native
 ordering and do not indicate a correspondence across layers or models.}
 \label{fig:head-heterogeneity}
\end{figure}

To examine how spectral structure varies across heads, we retain the
layer and head identities in the real-attention rank proxies and the
real-versus-canonical $W_1$ comparison. Figure~\ref{fig:head-heterogeneity}
shows each head's median over the same 12 GPT-2 documents at length
1,024 and 20 ViT images at length 577. Within each layer, the ratio of
the largest to the smallest head-specific median stable rank measures
the spread across heads. Its median across layers is $4.22$ for GPT-2
and $4.73$ for ViT; the corresponding ratios for $W_1$ are $3.19$ and
$1.49$ (Table~\ref{tab:head-heterogeneity}).

The head-resolved view reveals substantial variation in spectral
concentration within both architectures. It motivates studying how
the dominant token subspaces of different heads combine through the
value and output projections. Appendix~\ref{app:model-scope} formulates
this question at the level of the multi-head output matrix.

\subsubsection{Head-resolved spectral summaries}

Figure~\ref{fig:head-heterogeneity} reaggregates the saved per-head
measurements at GPT-2 length 1,024 and ViT length 577, using the same
12 documents and 20 images described above. Each model has 144
layer--head cells, with 12 or 20 observations per cell. With $x,l,h$
indexing inputs, layers, and heads, define for a diagnostic $q$
\[
m_{l,h}^{(q)}=\operatorname{median}_{x}q_{x,l,h},
\qquad
R_l^{(q)}=\frac{\max_h m_{l,h}^{(q)}}{\min_h m_{l,h}^{(q)}}.
\]
Here $q$ is the real-attention stable rank, real-attention $\erank_1$,
or the paired real-versus-canonical $W_1$. The heatmaps display
$m_{l,h}^{(q)}$, and Table~\ref{tab:head-heterogeneity} reports
$\operatorname{median}_l R_l^{(q)}$. All denominators are positive in
these data. Aggregation occurs after computing the diagnostics of each
sample--layer--head matrix; no attention matrices or singular vectors
are pooled across heads. These are descriptive summaries of the selected
inputs and layers. The analysis uses existing measurements and requires
no additional model inference.

\begin{table}[htbp]
 \centering
 \caption{Median across layers of the within-layer maximum-to-minimum
 ratio of head-specific median diagnostics.}
 \label{tab:head-heterogeneity}
 \begin{tabular}{lccc}
 \toprule
 Model & Stable rank & $\erank_1$ & $W_1$ \\
 \midrule
 GPT-2 ($\ell=1{,}024$) & 4.22 & 6.57 & 3.19 \\
 ViT ($\ell=577$) & 4.73 & 7.27 & 1.49 \\
 \bottomrule
 \end{tabular}
\end{table}

\clearpage
\section*{NeurIPS Paper Checklist}

\begin{enumerate}

\item {\bf Claims}
    \item[] Question: Do the main claims made in the abstract and introduction accurately reflect the paper's contributions and scope?
    \item[] Answer: \answerYes{} 
    \item[] Justification: The abstract and Introduction describe the canonical diagonal score model, spike--bulk separation, spectral degeneration, and the stable-rank contrast between bidirectional and causal attention. These claims are supported by Assumptions~\ref{ass:diagonal-in-law}--\ref{ass:spectral-separation}, Theorems~\ref{thm:singular-collapse}--\ref{thm:causal_singular_values}, and the rank-proxy results in Appendix~\ref{app:rank-proxies}. Section~\ref{experiment} provides empirical diagnostics on pretrained ViT/Imagenette and GPT-2/Wikitext-103.

\item {\bf Limitations}
    \item[] Question: Does the paper discuss the limitations of the work performed by the authors?
    \item[] Answer: \answerYes{} 
    \item[] Justification: Appendix~\ref{app:model-scope} discusses the canonical model's scope, orientation dependence, causal ordering, and the interpretation of the empirical diagnostics. The theoretical assumptions are explicit, and Section~\ref{experiment} and Appendix~\ref{app:additional-experimentation-details} describe the empirical scope of the pretrained ViT and GPT-2 experiments on Imagenette and Wikitext-103.

\item {\bf Theory assumptions and proofs}
    \item[] Question: For each theoretical result, does the paper provide the full set of assumptions and a complete (and correct) proof?
    \item[] Answer: \answerYes{} 
    \item[] Justification: Assumptions~\ref{ass:diagonal-in-law} and~\ref{ass:spectral-separation} specify the canonical model and separation conditions; additional conditions, including bounded bulk scores, are stated in the relevant theorems. The numbered results are cross-referenced, with complete spectral proofs in Appendices~\ref{app:proof-thm:singular-collapse}--\ref{app:proof-thm:causal_singular_values}, rank-proxy proofs in Appendix~\ref{app:rank-proxies}, and auxiliary matrix-norm proofs in Appendix~\ref{app:auxiliary-norm-bounds}.

    \item {\bf Experimental result reproducibility}
    \item[] Question: Does the paper fully disclose all the information needed to reproduce the main experimental results of the paper to the extent that it affects the main claims and/or conclusions of the paper (regardless of whether the code and data are provided or not)?
    \item[] Answer: \answerYes{} 
    \item[] Justification: Section~\ref{experiment} specifies the models, datasets, validation splits, sequence lengths, spectral metrics, and rank definitions. Appendix~\ref{app:additional-experimentation-details} provides sampling, SVD, angular-distance, and aggregation details, including sample counts, the seed for the matched attention comparison, and the interpretation of quantile bands.

\item {\bf Open access to data and code}
    \item[] Question: Does the paper provide open access to the data and code, with sufficient instructions to faithfully reproduce the main experimental results, as described in supplemental material?
    \item[] Answer: \answerYes{} 
    \item[] Justification: The experiments use public Imagenette and Wikitext-103 data and pretrained ViT and GPT-2 models. The repository at \url{https://github.com/Mengda-Li/attention} provides analysis code, saved figure inputs, environment and data-preparation instructions, and figure-to-script mappings. Its reproduction guide distinguishes rendering from supplied CSV or archived figure inputs from collecting model measurements and records the available collection workflows and their limitations; Appendix~\ref{app:additional-experimentation-details} links these materials.

\item {\bf Experimental setting/details}
    \item[] Question: Does the paper specify all the training and test details (e.g., data splits, hyperparameters, how they were chosen, type of optimizer) necessary to understand the results?
    \item[] Answer: \answerYes{} 
    \item[] Justification: The paper evaluates pretrained ViT-B/16 (SWAG) on Imagenette validation images and GPT-2 on Wikitext-103 validation text, without training new models. Section~\ref{experiment} and Appendix~\ref{app:additional-experimentation-details} specify the exact checkpoints, input dimensions, sequence lengths, sampling and seed, float64 attention SVD, score factorization, angular distances, and stable/effective-rank computations.

\item {\bf Experiment statistical significance}
    \item[] Question: Does the paper report error bars suitably and correctly defined or other appropriate information about the statistical significance of the experiments?
    \item[] Answer: \answerYes{} 
    \item[] Justification: Figures~\ref{fig:wikitext-distance-metrics}, \ref{fig:score-rank-wikitext}, and~\ref{fig:gpt2-real-canonical-w1} report means or medians with quantile bands; Figures~\ref{fig:vit-layer-distance-boxplot} and~\ref{fig:vit-real-canonical-w1} use layer-wise box plots. The captions and Appendix~\ref{app:real-canonical-comparison} define these summaries. They describe empirical variability across sampled observations, including dependent heads and nested text prefixes in the matched comparison, rather than confidence intervals.

\item {\bf Experiments compute resources}
    \item[] Question: For each experiment, does the paper provide sufficient information on the computer resources (type of compute workers, memory, time of execution) needed to reproduce the experiments?
    \item[] Answer: \answerYes{} 
    \item[] Justification: Appendix~\ref{app:additional-experimentation-details} identifies the compute used for each experiment group: hidden-state angular diagnostics and GPT-2 score-rank summaries used an AMD EPYC 9554 dual-socket CPU server with 64 cores per socket and 126 GB RAM, without GPU acceleration, for approximately 24 hours in total. Real-versus-canonical comparisons and score-level angular and spectral diagnostics used a MacBook Pro with an M1 Pro chip and 16 GB unified memory for approximately two hours in total. The reported timings are totals for each group.
    
\item {\bf Code of ethics}
    \item[] Question: Does the research conducted in the paper conform, in every respect, with the NeurIPS Code of Ethics \url{https://neurips.cc/public/EthicsGuidelines}?
    \item[] Answer: \answerYes{} 
    \item[] Justification: The work is theoretical and diagnostic, uses public benchmark data and pretrained models, and does not involve human subjects, private data, or deployed systems. The paper does not identify any confidential or sensitive data sources.

\item {\bf Broader impacts}
    \item[] Question: Does the paper discuss both potential positive societal impacts and negative societal impacts of the work performed?
    \item[] Answer: \answerNA{} 
    \item[] Justification: The paper is foundational theory and empirical diagnosis of attention spectra, and it does not introduce a deployable model, dataset, or capability with a direct path to societal deployment. Any downstream impacts would be indirect through broader understanding of Transformer mechanisms.
    
\item {\bf Safeguards}
    \item[] Question: Does the paper describe safeguards that have been put in place for responsible release of data or models that have a high risk for misuse (e.g., pre-trained language models, image generators, or scraped datasets)?
    \item[] Answer: \answerNA{} 
    \item[] Justification: The paper does not release a new high-risk model, image generator, language model, or scraped dataset. It releases analysis/plotting code for reproducing figures, so additional model-release safeguards are not applicable.

\item {\bf Licenses for existing assets}
    \item[] Question: Are the creators or original owners of assets (e.g., code, data, models), used in the paper, properly credited and are the license and terms of use explicitly mentioned and properly respected?
    \item[] Answer: \answerYes{} 
    \item[] Justification: Appendix~\ref{app:asset-licenses} identifies the existing datasets, code libraries, and model weights used in the experiments and lists the corresponding licenses or access terms, including ImageNet/Imagenette, Wikitext-103, Hugging Face Transformers, GPT-2 weights, TorchVision, and SWAG-derived ViT weights. The paper cites the creators or source repositories and uses these assets only for non-commercial research evaluation and spectral diagnostics.

\item {\bf New assets}
    \item[] Question: Are new assets introduced in the paper well documented and is the documentation provided alongside the assets?
    \item[] Answer: \answerYes{} 
    \item[] Justification: The released analysis and plotting code is documented alongside the code with its inputs, dependencies, execution commands, output definitions, and reproduction limitations. The authors' code is released under the MIT license; existing data and models retain their original terms. Appendix~\ref{app:additional-experimentation-details} links the repository and reproduction guides. No new dataset, benchmark, or trained model is introduced.

\item {\bf Crowdsourcing and research with human subjects}
    \item[] Question: For crowdsourcing experiments and research with human subjects, does the paper include the full text of instructions given to participants and screenshots, if applicable, as well as details about compensation (if any)? 
    \item[] Answer: \answerNA{} 
    \item[] Justification: The work does not involve crowdsourcing, participant studies, data annotation by human subjects, or user-facing experiments.

\item {\bf Institutional review board (IRB) approvals or equivalent for research with human subjects}
    \item[] Question: Does the paper describe potential risks incurred by study participants, whether such risks were disclosed to the subjects, and whether Institutional Review Board (IRB) approvals (or an equivalent approval/review based on the requirements of your country or institution) were obtained?
    \item[] Answer: \answerNA{} 
    \item[] Justification: No human-subject research is conducted, so IRB approval or equivalent human-subject review is not applicable.

\item {\bf Declaration of LLM usage}
    \item[] Question: Does the paper describe the usage of LLMs if it is an important, original, or non-standard component of the core methods in this research? Note that if the LLM is used only for writing, editing, or formatting purposes and does \emph{not} impact the core methodology, scientific rigor, or originality of the research, declaration is not required.
    \item[] Answer: \answerYes{}
    \item[] Justification: LLM assistance was used in revising the manuscript and checking mathematical arguments. The resulting statements are accompanied by explicit assumptions and proofs; LLM output is not treated as mathematical evidence.

\end{enumerate}

\end{document}